\documentclass[10pt,twocolumn,letterpaper]{article}

\usepackage{cvpr}
\usepackage{times}
\usepackage{epsfig}
\usepackage{graphicx}
\usepackage{amsmath}
\usepackage{amssymb}
\usepackage{cite}
\usepackage{xcolor}
\usepackage{enumitem}
\usepackage{subfigure}
\usepackage{textcomp}
\usepackage{authblk}
\usepackage[marginal]{footmisc}

\newtheorem{theorem}{Theorem}

\cvprfinalcopy 

\def\cvprPaperID{10} 

\ifcvprfinal\fi
\begin{document}

\title{Implicit Euler ODE Networks for Single-Image Dehazing}


\author{Jiawei Shen$^{\ast1}$, Zhuoyan Li$^{\ast1}$, Lei Yu$^{\dag1}$, Gui-Song Xia$^2$, Wen Yang$^1$\\
$^1$School of Electronic and Information, Wuhan University\\
$^2$School of Computer Science, Wuhan University\\
Wuhan 430072, China\\
{\tt\small \{shenjiawei0116, xzflzy\}@gmail.com, \{ly.wd, guisong.xia, yangwen\}@whu.edu.cn}
}

\maketitle
\thispagestyle{empty}

\footnote{\noindent
$\ast$ The first two authors contributed equally and should be regarded as co-first authors.\\
$\dag$ Corresponding author.
}

\begin{abstract}
Deep convolutional neural networks (CNN) have been applied for image dehazing tasks, where the residual network (ResNet) is often adopted as the basic component to avoid the vanishing gradient problem. Recently, many works indicate that the ResNet can be considered as the explicit Euler forward approximation of an ordinary differential equation (ODE). In this paper, we extend the explicit forward approximation to the implicit backward counterpart, which can be realized via a recursive neural network, named IM-block. Given that, we propose an efficient end-to-end multi-level implicit network (MI-Net) for the single image dehazing problem. Moreover, multi-level fusing (MLF) mechanism and residual channel attention block (RCA-block) are adopted to boost performance of our network. Experiments on several dehazing benchmark datasets demonstrate that our method outperforms existing methods and achieves the state-of-the-art performance.
\end{abstract}

\section{Introduction}
Images captured from the harsh environment are often hazy and exhibit a reduced visibility of scenery \cite{he2010single} with the loss of contrast, color fidelity and edge information. Such degradation of images may greatly decrease the accuracy and robustness for the subsequent high-level computer vision tasks \cite{zhang2018densely,li2018single}. Given that corrupted images, many recent work on semantic foggy scene understanding \cite{sakaridis2018semantic,dai2019curriculum} are applied as a preprocessing step for high-level semantic tasks.  which makes the single-image dehazing an urgent but challenging task \cite{cheng2018semantic,galdran2018duality}. Existing approaches, either exploiting the physical prior \cite{he2010single,pei2012nighttime,berman2017air,sulami2014automatic} or learning the inverse mapping of degradation with large datasets \cite{cai2016dehazenet,li2017aod,ren2018gated,dong2020fd}, have been extensively investigated.



For CNN-based image dehazing methods, residual block (Resblock) \cite{he2016deep} is widely applied \cite{liu2019griddehazenet,ren2018gated,zhang2018densely,qin2019ffa}. Zhang et al.\cite{zhang2018densely} introduce a densely connected pyramid network with residual block, Liu et al.\cite{liu2019griddehazenet} has leveraged the residual dense block in their GRID-Net and Qin et al.\cite{qin2019ffa} develop a residual attention block in their FFA-Net. Overall, residual block is not only widely applied in high-level tasks but also low-level tasks as image dehazing. On the other hand, we notice that many recent works relate ResNet with explicit Euler forward approximation of ODE.

 \begin{figure}[t]
\centering
\subfigure[Hazy Image]{
\includegraphics[width=4cm]{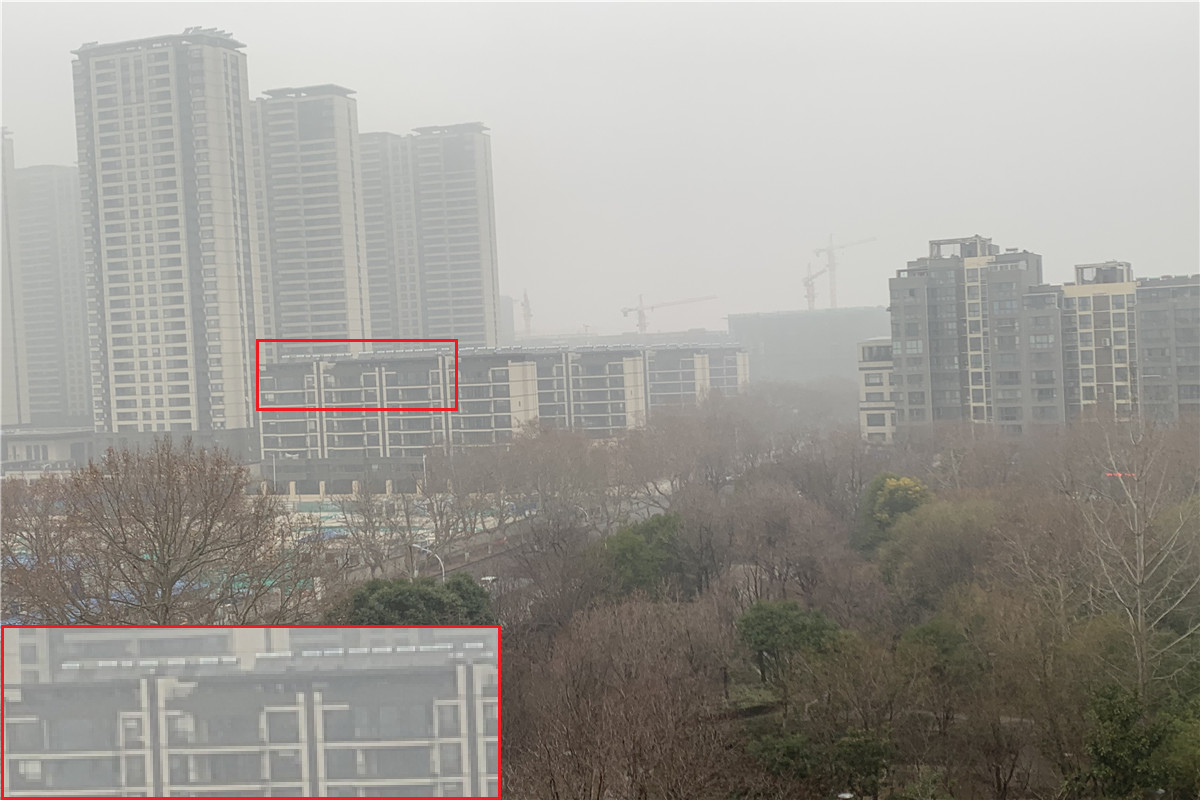}
}%
\subfigure[Ours ($0.628M$)]{
\includegraphics[width=4cm]{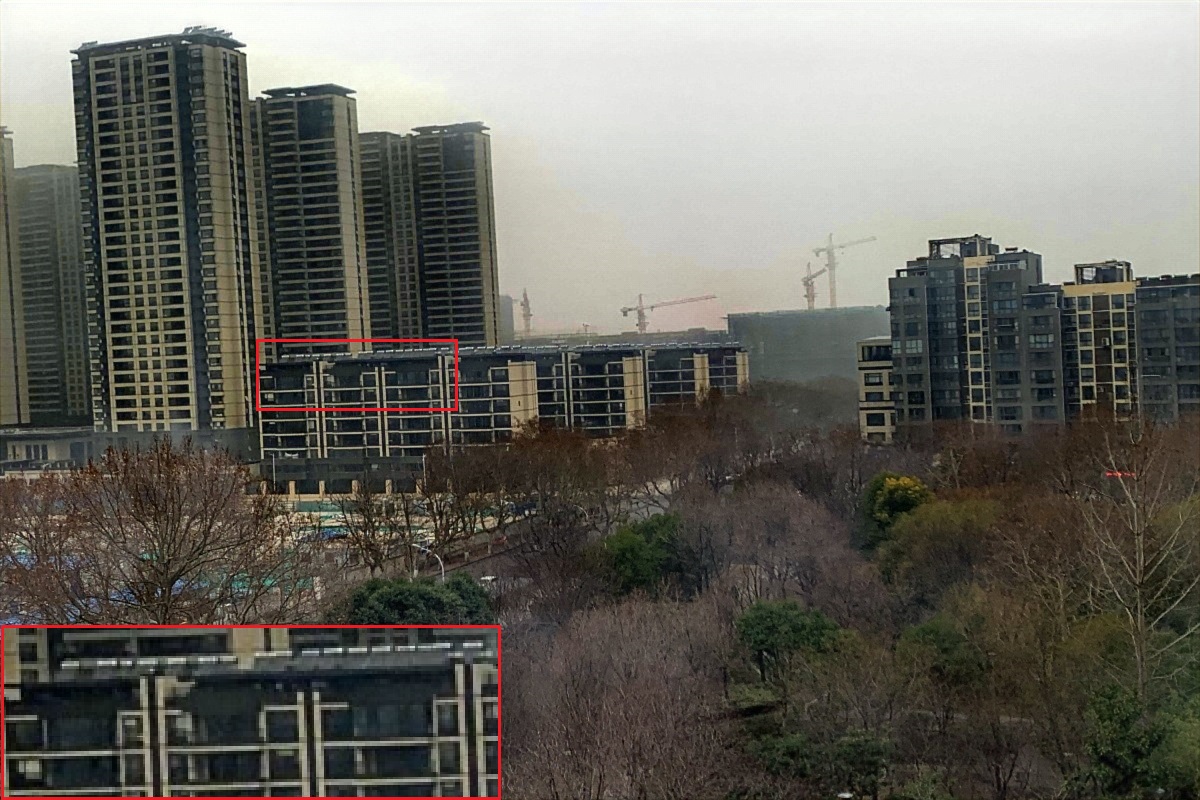}
}
\quad
\subfigure[FFA-Net ($4.46M$)]{
\includegraphics[width=4cm]{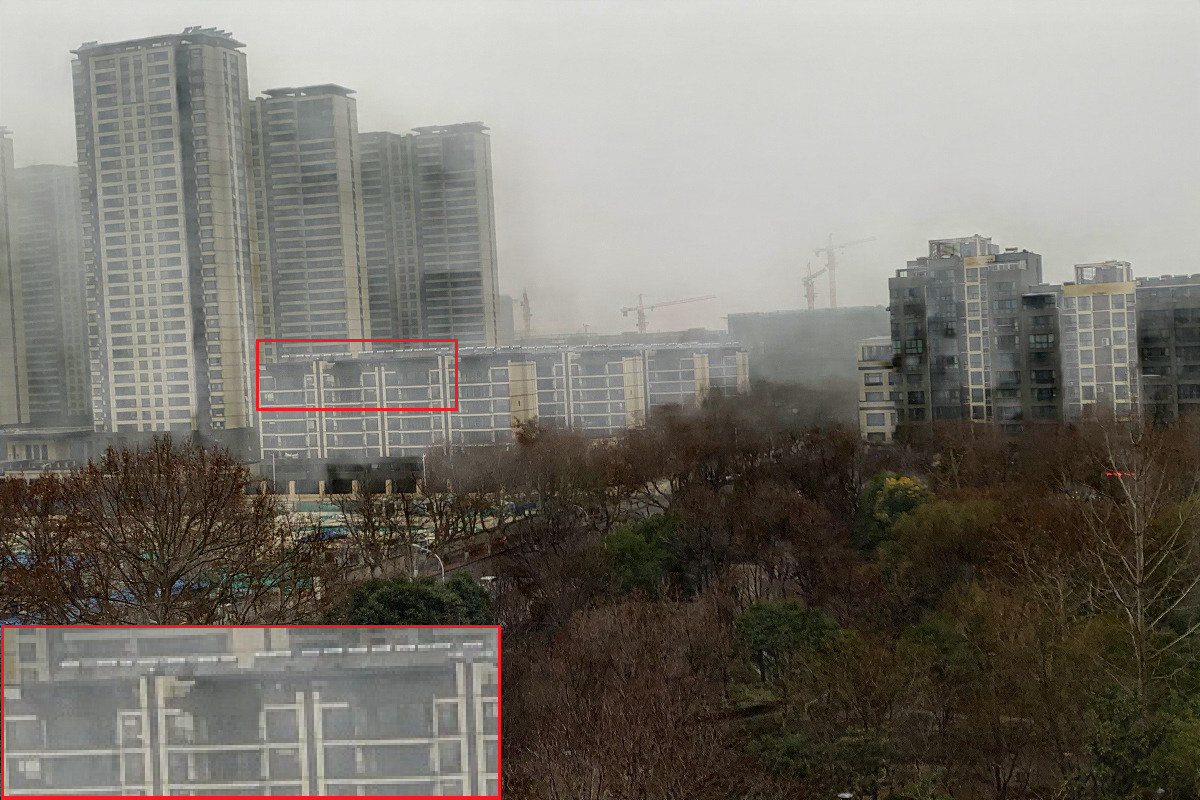}
}%
\subfigure[GRID-Net ($0.958M$)]{
\includegraphics[width=4cm]{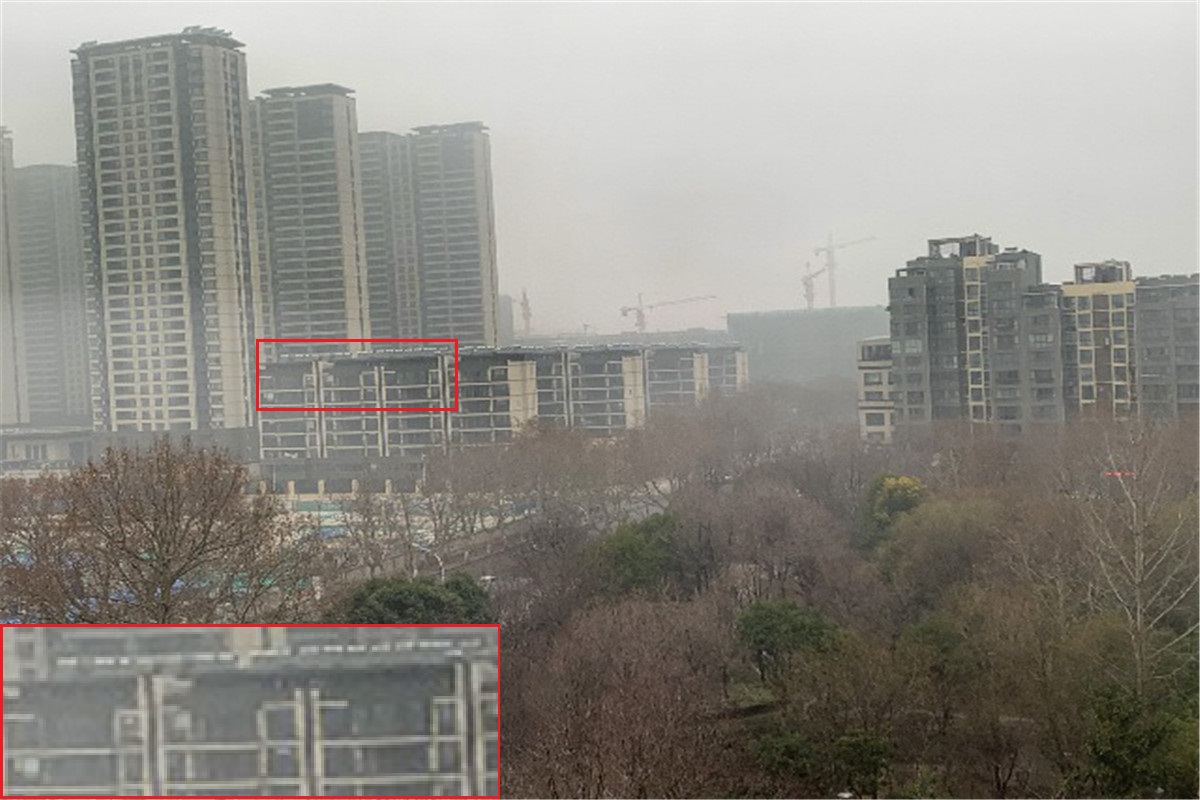}
}
\caption{Visual comparison of our method and other state-of-the-art works, FFA\cite{qin2019ffa} and GRID-Net\cite{liu2019griddehazenet}. The numbers in brackets indicate the size of the networks.}
\label{fig1}
\end{figure}

Researching CNNs' theoretical properties and behavior has drawn considerable attention from the perspective of ODE \cite{ruseckas2019differential,thorpe2018deep,chen2018neural}. Chen et al.\cite{chen2018neural} introduce the relation between CNN and ODE. Ruseckas et al.\cite{ruseckas2019differential} prove that not only residual networks, but also feedforward neural networks with small nonlinearities can be related to the differential equations. Thorpe et al. \cite{thorpe2018deep} propose that residual neural network model is a discretization of an explicit Euler ODE and the deep-layer limit coincides with a parameter estimation problem for a nonlinear ordinary differential equation. These works are all closely associated with ResNet, which is an explicit Euler scheme. However, focusing on parameters convergence and system stability, implicit Euler has been proved to be better than explicit one since implicit Euler is unconditionally stable\cite{incc1998comparison}. Moreover, considering that the original objective of Resblock applied in deep neural network is to overcome vanishing gradient problem, it's suboptimal to directly apply Resblock in low-level tasks as image dehazing.

\begin{figure*}[ht]
    \centering
    \includegraphics[width=0.9573\textwidth]{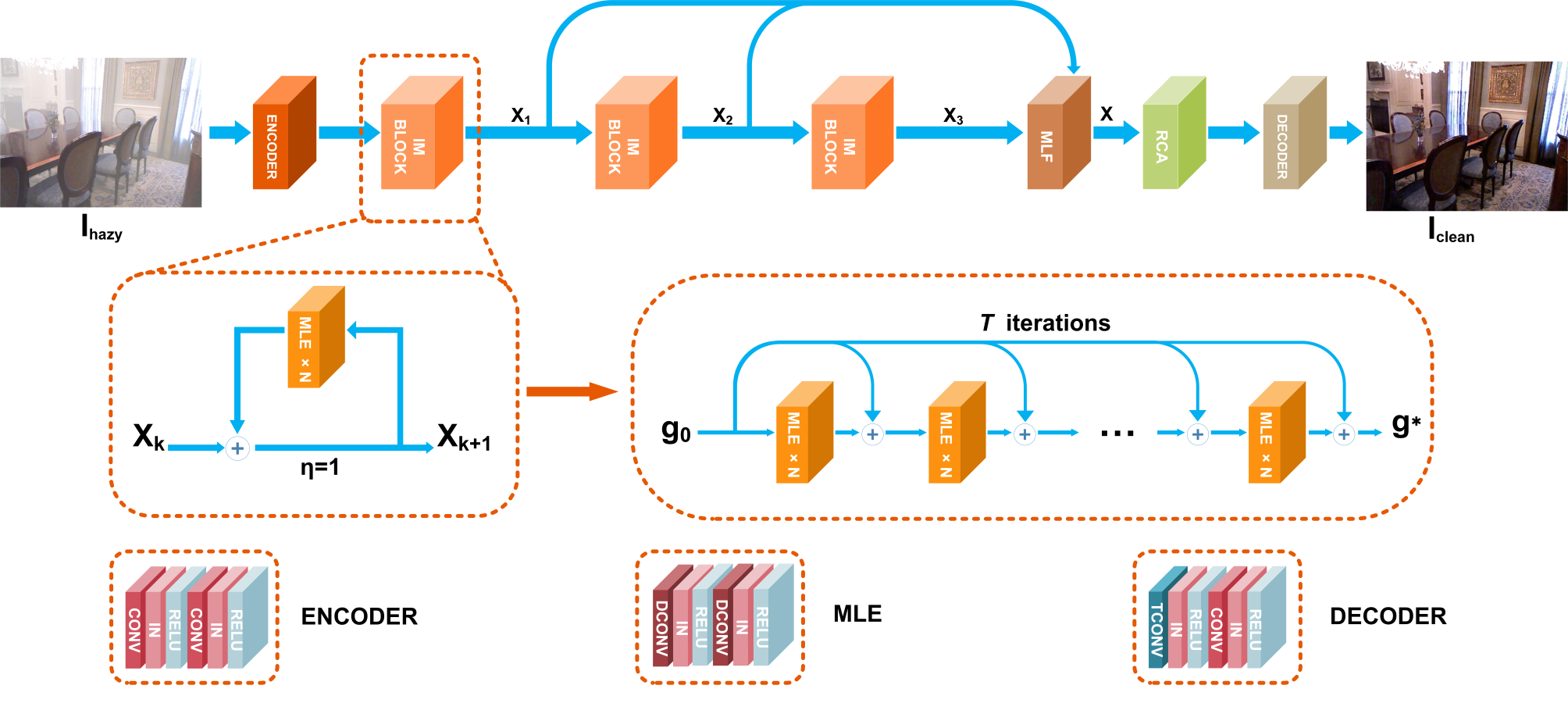}
    \caption{The architecture of multi-level implicit network (MI-Net). IM-block is illustrated in dashed box. \textrm{IN} and \textrm{DCONV} in MLE denotes instance normalization and dilated convolution respectively. $\times N$ in MLE represents the dilated rate and the dilated rates are different for each IM-block. \textrm{TCONV} in \textrm{DECODER} represents fractionally-strided convolution. MLF and RCA-block will be illustrated in proposed methods section. }
    \label{MINET}
\end{figure*}

In this work, we propose a novel efficient network, multi-level implicit network (MI-Net), for single image dehazing. Specially, the structure of MI-Net is shown in Fig.~\ref{MINET}. We introduce an implicit block (IM-block) combining the merits of implicit Euler scheme and CNN to realize implicit discretization of an implicit Euler ODE issue. In network's architecture, we cascade three IM blocks to learn the mapping relation between hazy image and clean image. Inspired from \cite{lin2017feature}, we integrate features from three IM blocks with different weight coefficients generated from MLF block. Noting that the fused feature would inevitably incorporate artifact, the fused feature is fed into RCA-block to suppress artifact of channels. As we shall see in the experiments, our method achieves the best balance between the performance and size.

 In summary, our contributions are as follows:
  \begin{itemize}[leftmargin = 8 mm,topsep = 0 pt]
    \item[$\bullet$] We propose a simple yet effective implicit scheme framework for single image dehazing, which simplifies the net significantly but boosts the dehazing performance compared to the residual framework.
    \item[$\bullet$] We propose a residual channel attention block based on attention mechanism to alleviate the artifact but retain abundant texture features.
    \item[$\bullet$] Experimental results on several widely used benchmark datasets show superior performance of our method compared to state-of-the-art methods, which verifies the effectiveness of our method.
 \end{itemize}
\section{Related Work}

\subsection{Image Dehazing}

\textbf{Prior-driven Methods:} The procedure of classical dehazing task is the reverse procedure of the atmospheric scattering model described as $I(p) = t(p)R(p) + A(1-t(p))$, where $I(p)$ is the hazy image, $ t(p)$ represents transmission map, $R(p)$ is the clean image, $p$ represents the pixel location and A is the global atmospheric light constant value. The traditional methods use assumptions which derive from the statistical characteristics of hazy image to compensate for the loss information in the atmospheric model. Tan et al. \cite{tan2008visibility} builds Markov random field cost function to obtain the clean image by maximizing the contrast of corrupted image ,assuming that the contrast of clean image is higher than that of hazy image and the smoothness of global atmospheric light. He et al. \cite{he2010single} proposes the dark channel prior, which is based on a statistical observation that at least one of color channels in the non-sky regions of a hazy image has a value close to zero, to estimate the transmission map. Fattal et al. \cite{fattal2008single} uses a color-line method line for dehazing based on observation that the color distribution of small patches of image in the RGB space is one-dimensional. Although these methods could handle the problem to some extent, the priors they based on would be invalid in some real scenes, which enables the methods struggle.

\textbf{CNN-based Methods:} Recently, many CNN-based methods have been applied in the dehazing by leveraging the advancement of powerful GPU and large-scale datasets. Early CNN-based methods still are based on global atmospheric scattering model and recover clean image by estimating the transmission map and global atmospheric light. Ren et al. \cite{ren2016single} design a coarse-to-fine multiscale network to estimate a refined transmission map. Cai et al. \cite{cai2016dehazenet} develop a Dehaze-Net embedded with feature layers for prediction of transmission map. However, the transmission map is susceptible to the noise and hence reduces the quality of dehazing performance. Therefore, end-to-end CNNs have been proposed to output a clean image from a hazy image directly without global atmospheric scattering model. Chen et al. \cite{chen2019gated} use an encoder-decoder net with smoothed dilated convolution in the net to alleviate the gridding artifacts. Zhang et al. \cite{zhang2018densely} propose a densely-connected pyramid densely network (DCPDN) to jointly learn clean image, atmospheric light and transmission map. Although the DCPDN improves the performance to some extent, it enlarges the size of model greatly. Deng et al. \cite{deng2019deep} develop a net that fuses the atmospheric scattering model with extracted haze together to improve the dehazing results. We note that most recent works still rely on deepening the net to improve the quality whatever the principles they based on. Being a low-level task, image dehazing depends more on low-level features compared to high-level features. We consider the dehazing network could be simplified into a low-level network.

\subsection{CNN from Ordinary Differential Equation}
Along with the development of CNN, many researches study networks' theoretical properties from the perspective of ordinary differential equation (ODE) \cite{david2018ordinary,hoops2016ordinary,bonin2017mathematical}. Weinan et al. \cite{weinan2017proposal} firstly view ResNet \cite{he2016deep} as an approximation to ODE, which exploits the possibility of using computational theory from ResNet's dynamic system. Similarly, Chang et al. \cite{chang2018reversible} connect ResNet with nonlinear ODE, and extend three reversible network architectures. Chen et al.\cite{chen2018neural} propose ODE-Net that combines ODE and CNN. Based on above, we speculate that network designed with certain theoretical basis has great potential for exploration. Thorpe et al.\cite{thorpe2018deep} considered that the deep layer limit coincides with a parameter estimation problem for nonlinear ODE \cite{david2018ordinary}. In reality, convergent parameters means stable system which is closely related to model performance. Haber et al. \cite{haber2017stable} also associated gradient explosion and disappearance about neural networks with the stability of discrete ODE and suggested that stable networks generalize better. In fact, although implicit Euler scheme has larger computational cost compared to explicit Euler scheme, implicit one allows greater step size and is more stable since implicit scheme is unconditionally stable. Moreover, for low-level task as image dehazing, the increased computational cost could be ignored. Considering these all factors, we adopt the implicit Euler scheme in CNN to determine the dehazing model.

\section{Proposed Methods}
As shown in Fig.~\ref{MINET}, we build an end-to-end network to establish mapping relation between hazy image and clean image. In this section, we illustrate the function of each component in MI-Net's architecture. Firstly, the hazy image input $I_{hazy}$ will be encoded from RGB space to feature space. The encoded features will be enhanced by three cascaded IM-block. Additionally, the features $x_1, x_2, x_3$ output from each IM-block will be fused in MLF block. MLF block is an operation block, three features are input into MLF then MLF learns the weight coefficient of each feature $x_i$, and the output of MLF is a fusion of three features combined with different weights. The fused feature is input to RCA-block to refine the feature. Lastly, the refined feature will be decoded into RGB space to get the clean image $I_{clean}$.

\begin{figure}[t]
    \centering
    \includegraphics[scale=.8]{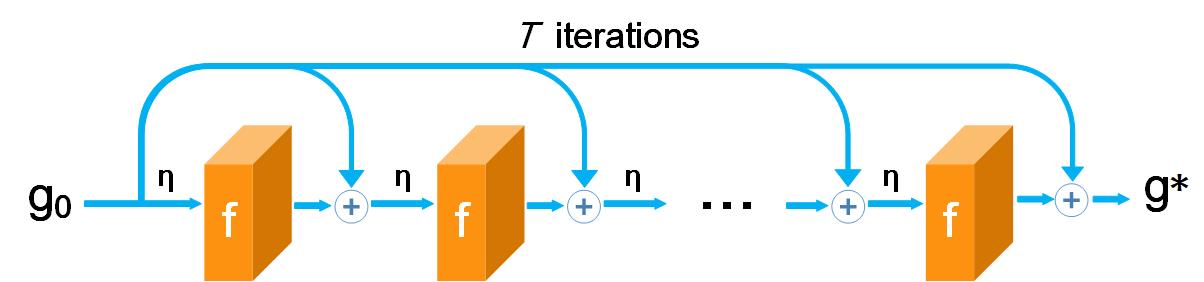}
    \caption{IM-block: Implicit scheme with specific function $f$.}
    \label{IMscheme}
\end{figure}

\subsection{From Resblock to IM-block}


Convolutional neural network (CNN) has been applied to tackle the single image dehazing problem. By increasing the depth of CNN, it is possible to improve the perceptual fields and increase the expressive ability, thus leading to better dehazing performance \cite{qin2019ffa,dong2020fd}.
Residual neural network (ResNet) \cite{he2016deep} has been widely used in both high-level and low-level tasks to overcome the vanishing gradient problem for deep neural networks. Recently, from the point view of ODE, Resblock can be considered as the explicit Euler forward discretization of the continuous-time ODE: $\dot{x}(t) = f(x(t))$ \cite{chen2018neural} and the relation between two consecutive layers can be expressed as
\begin{equation}\label{eq:resnet}
\mbox{\bf Resblock:} \quad 	x_{k+1} = x_k + \eta f(x_{k}),
\end{equation}
where $\eta$ denotes the discretization step.


On the other hand, implicit Euler backward scheme  \cite{incc1998comparison} has been proven to be more stable and accurate compared to the explicit counterpart \eqref{eq:resnet}. From this point, the implicit scheme that bridges two consecutive layers can be written as
 \begin{equation}\label{eq:im}
    \mbox{\bf IM-block:} \quad x_{k+1} = x_k + \eta f(x_{k+1}).
 \end{equation}

However, the above implicit algebraic equation cannot be solved analytically in general, and Newton iteration method is generally used to find $x_{k+1}$. Letting $g_0\triangleq x_k$, the following iterations can be applied to approximate the solution:
 \begin{equation}\label{eq:g_iter}
    g_{\kappa+1} = g_0 + \eta f(g_\kappa) 	
 \end{equation}
and $x_{k+1} = g^*$, with $g^*$ the equilibrium point of \eqref{eq:g_iter}, i.e.
 \begin{equation}\label{eq:g_star}
    g^* = g_0 + \eta f(g^*)
 \end{equation}

Then we have the following theorem,
\begin{theorem}
	Let $\partial_x f(x) \in \mathbb{R}^{n\times n}$ be the partial derivative of the vector field $f(x)$ at $x$, and note $\lambda_i$ as its eigenvalue for $1\leq i\leq n$. Then \eqref{eq:g_iter} is stable if
	\begin{equation}\label{eq:condition}
		|\lambda_i|< 1/\eta		
	\end{equation}
 for all $1\leq i\leq n$.
\end{theorem}
The proof is straightforward. The above theorem implies that the vector field $f(x)$ should be well defined to guarantee the convergence of \eqref{eq:g_iter}. Thus in our proposed MI-Net, the instance normalization \cite{ulyanov2016instance} \textrm{IN} is exploited to ensure \eqref{eq:condition}, as shown in Fig.~\ref{MINET}. On the other hand, iterative approximation usually uses a large finite number as infinite iterations and we illustrate the unfolded form of \eqref{eq:g_iter}, named IM-block in Fig.~\ref{IMscheme}, which is exploited as the basic block instead of ResNets to construct our proposed MI-Net.



\begin{figure}[]\setcounter{subfigure}{0}
    \centering%
    \begin{minipage}{0.31\linewidth}
        \centering
        \includegraphics[width=1in]{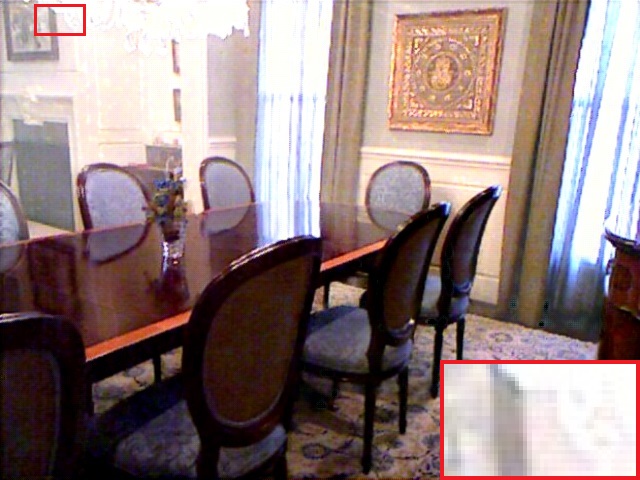}
        \hspace{.05in}
        \centerline{$I_{x_1}$}
    \end{minipage}
    \begin{minipage}{0.31\linewidth}
        \centering
        \includegraphics[width=1in]{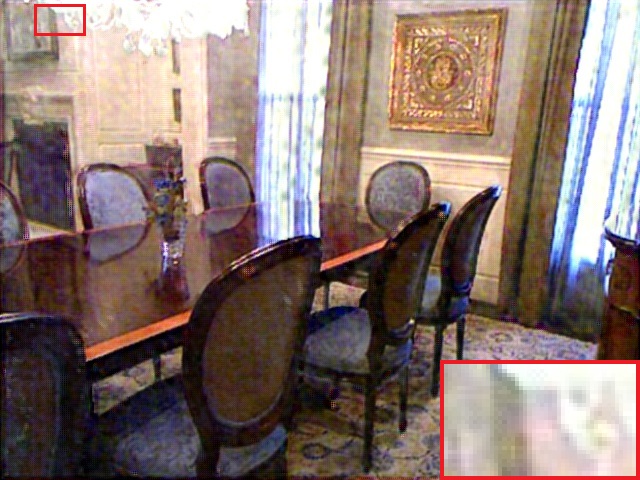}
        \hspace{.05in}
        \centerline{$I_{x_2}$}
    \end{minipage}
    \begin{minipage}{0.31\linewidth}
        \centering
        \includegraphics[width=1in]{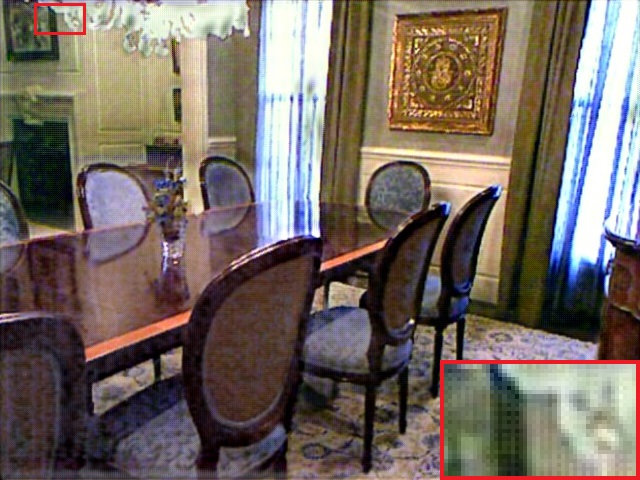}
        \hspace{.05in}
        \centerline{$I_{x_3}$}
    \end{minipage}

\centering
\caption{The visual comparison of $I_{x_1}, I_{x_2}, I_{x_3}$. $I_{x_1}$ has clear detail information but still has hazy area. $I_{x_3}$ has gridding effect and color distortion area, i.e. the edge of ceiling lamp. Please zoom in for a better illustration.}
\label{x}
\end{figure}

\begin{table*}[t]
\centering
\caption{Quantitative comparisons of image dehazing on SOTS dataset from RESIDE, TestA and MiddleBury.}
\label{performance}
\setlength{\tabcolsep}{2.5mm}{
\begin{tabular}{ccccccccc}
\hline
\multicolumn{1}{c|}{PSNR (dB)}  & \begin{tabular}[c]{@{}c@{}} DCP \\ \cite{he2010single}   \end{tabular}  & \begin{tabular}[c]{@{}c@{}}AOD-Net \\ \cite{li2017aod}\end{tabular} & \begin{tabular}[c]{@{}c@{}}DCPDN\\ \cite{zhang2018densely} \end{tabular} & \begin{tabular}[c]{@{}c@{}}GFN\\ \cite{ren2018gated} \end{tabular} & \begin{tabular}[c]{@{}c@{}}GRID-Net\\ \cite{liu2019griddehazenet}\end{tabular} & \begin{tabular}[c]{@{}c@{}}FFA-Net\\ \cite{qin2019ffa}\end{tabular} & \multicolumn{1}{c|}{\begin{tabular}[c]{@{}c@{}}FD-GAN\\ \cite{dong2020fd} \end{tabular}} & OURS   \\ \hline
\multicolumn{1}{c|}{SOTS}  & 16.62  & 20.86    & 28.13     & 21.14     & 32.16        & \textbf{36.12}     & \multicolumn{1}{c|}{23.15}  & 35.51  \\
\multicolumn{1}{c|}{TestA}  & 13.91  & 20.46    & 23.27    & 20.02    & 22.33       &19.96     & \multicolumn{1}{c|}{18.82}  & \textbf{29.45}  \\
\multicolumn{1}{c|}{MiddleBury} & 11.94  & 13.94   & 14.31     & 14.01     & 12.83         & 13.76     & \multicolumn{1}{c|}{14.63}   & \textbf{17.44}  \\ \hline
                           &        &                                                        &                                                          &                                                       &                                                         &                                                        &                                                                            &        \\ \hline
\multicolumn{1}{c|}{SSIM}  & \begin{tabular}[c]{@{}c@{}} DCP \\ \cite{he2010single}   \end{tabular} & \begin{tabular}[c]{@{}c@{}}AOD-Net\\ \cite{li2017aod})\end{tabular} & \begin{tabular}[c]{@{}c@{}}DCPDN\\ \cite{zhang2018densely} \end{tabular} & \begin{tabular}[c]{@{}c@{}}GFN\\  \cite{ren2018gated}\end{tabular} & \begin{tabular}[c]{@{}c@{}}GRID-Net\\ \cite{liu2019griddehazenet} \end{tabular} & \begin{tabular}[c]{@{}c@{}}FFA-Net \\ \cite{qin2019ffa}\end{tabular} & \multicolumn{1}{c|}{\begin{tabular}[c]{@{}c@{}}FD-GAN\\ \cite{dong2020fd}\end{tabular}} & OURS   \\ \hline
\multicolumn{1}{c|}{SOTS}  & 0.8179 & 0.8788    & 0.9592   & 0.8500    & 0.9836      & \textbf{0.9886}   & \multicolumn{1}{c|}{0.9207}    & 0.9841 \\
\multicolumn{1}{c|}{TestA}  & 0.8642 & 0.8379    & 0.8398  & 0.8160     & 0.9123      &0.7715   & \multicolumn{1}{c|}{0.8614}    & \textbf{0.9394} \\
\multicolumn{1}{c|}{MiddleBury} & 0.7620 & 0.7426   & 0.7643   & 0.7545      & 0.6755        & 0.6983  & \multicolumn{1}{c|}{0.7812}  &\textbf{0.8465} \\ \hline
\end{tabular}}
\end{table*}


One can easily find that the ResNet \eqref{eq:resnet} actually equals to IM-block when it contains only one recursion \eqref{eq:g_iter} with $\kappa=0$. While the implementation of IM-block shown in Fig.~\ref{IMscheme} exhibits many recursions with shared weights for different layers. Theoretically, IM-block is equivalent to one layer implicit Euler approximation \eqref{eq:im}. In some sense, the IM-block could possess the perceptual field as large as the whole image but with a very shallow depth, and thus could be a more efficient structure than ResNet to capture low-level features. This property is favorable for low-level tasks \cite{sogaard2016deep} and motivates us to turn to IM-block for the image dehazing problem.

Consequently, we stack three IM-blocks as shown in Fig.~\ref{MINET} to capture features with different level of scales. To achieve this target, the multi-level feature fusion strategy is exploited in next subsection.

\subsection{Multi-Level Feature Fusion}
 Dilated convolutions \cite{yu2015multi} support exponential expansion of the receptive field without loss of resolution or coverage, which can be utilized in IM-block to further enlarge the receptive field in MI-Net. Thus we adopt a dilated convolutional block to serve as $f$ in \eqref{eq:g_iter}. To overcome the gridding effect of dilated convolution, Wang et al.\cite{wang2018understanding} proposed hybrid dilated convolution (HDC). Therefore, we adopt HDC structure that with coprime dilated rates, e.g. 1,2,5, to gain the best result for each IM-block. In order to help the convergence of \eqref{eq:g_iter}, we adopt instance norm layers instead of setting the discretization step $\eta$ a small value \cite{lim2017enhanced}. And the layers with dilated convolutions and instance normalizations are grouped as the multi-level extraction (MLE) block, as shown in Fig.~\ref{MINET}, where the \textrm{DCONV} represents dilated convolutions and the \textrm{IN} layer denotes instance normalization \cite{ulyanov2016instance}.


However, the HDC structure can only suppress the gridding effects, color distortion still exists in the third IM-block. We visualize the difference among features of each IM-block, we extract the output features of IM-blocks $x_1, x_2, x_3$ with $x_i \in \mathbb R^{c\times h\times w}$, where $c$ represents channel number, $h$ and $w$ represents image's height and width of input images. Without being fed into MLF, the features are directly input into \textrm{DECODER} block to transform features into images, i.e $I_{x_1}, I_{x_2}, I_{x_3}$. The corresponding reconstructions are plotted in Fig.~\ref{x}. Note that feature from MI-Net at first IM block contains detail information but the non-haze details of image reconstructed from this feature would still be corrupted with haze severely. Third IM block of MI-Net has a larger receptive field but ignore background details.
Hence, we adopt the attention based multi-level fusion block (MLF), and this block leverages convolutional layer to obtain a weight coefficient matrix $W\in \mathbb{R}^{3\times h\times w}$ of each pixel of feature $x \in \mathbb{R}^{c\times h\times w}$. In this case, the gridding effect area, the hazy area and the color distortion area will have a small weight coefficient but the weight coefficient of clean area will be larger.

Specially, as shown in Fig.~\ref{MINET}, MLF block obtains output features of three IM-blocks in three different levels $x_1, x_2, x_3$, the concatenation of three features $X = [x_1, x_2, x_3]$ where $ X\in \mathbb{R}^{3c\times h\times w}$ is fed into MLF layer to output weight coefficient $W= [W_1,W_2,W_3]$ where $W\in \mathbb{R}^{3\times h\times w}$ :
\begin{align}
    W = \textrm{MLF}(X)
\end{align}
where MLF is a group convolutional layer.

Finally, we generate fused feature $x$ by element-wise multiplying $\circ$, Hadamard product, $x_1, x_2, x_3$ with weight coefficient $W_1, W_2, W_3$ linearly in each $c$ channel:
\begin{align}
    x = W_1\circ x_1 + W_2\circ x_2 + W_3\circ x_3
\end{align}
Then the fused feature $x$ will be fed into RCA-block.

\begin{figure}[t]
    \centering
    \includegraphics[scale=1.3]{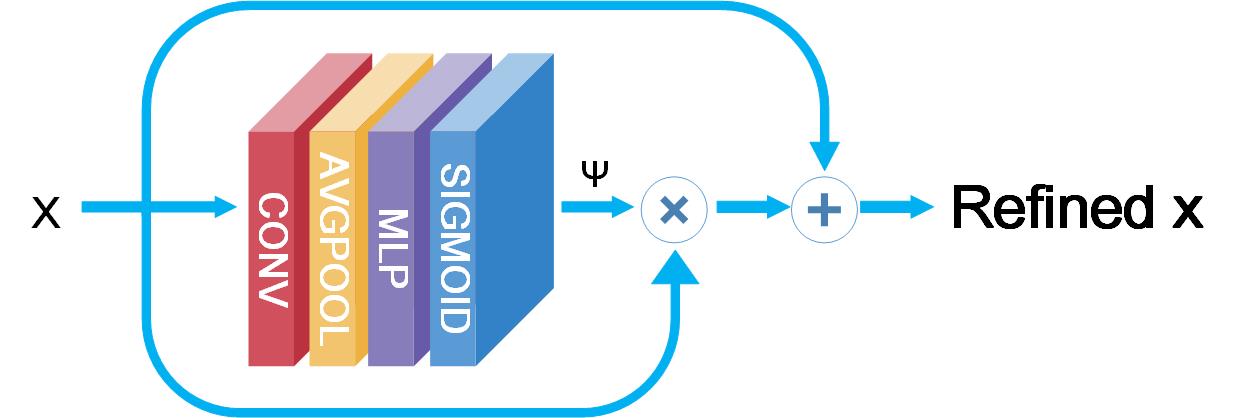}
    \caption{The schematic structure of Residual Channel Attention block.}
    \label{RCA}
\end{figure}

\begin{figure*}[t]\setcounter{subfigure}{0}
    \centering%
        \subfigure[HAZY]{
        \begin{minipage}[t]{0.135\linewidth}
            \centering
            \includegraphics[width=0.95in]{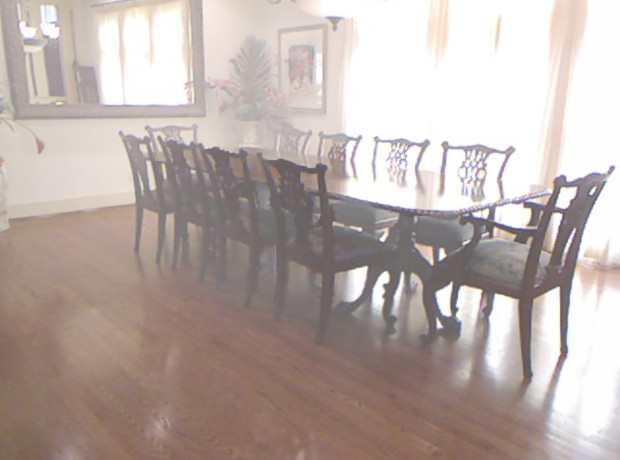}
            \centerline{-}
            \hspace{.05in}

            \includegraphics[width=0.95in]{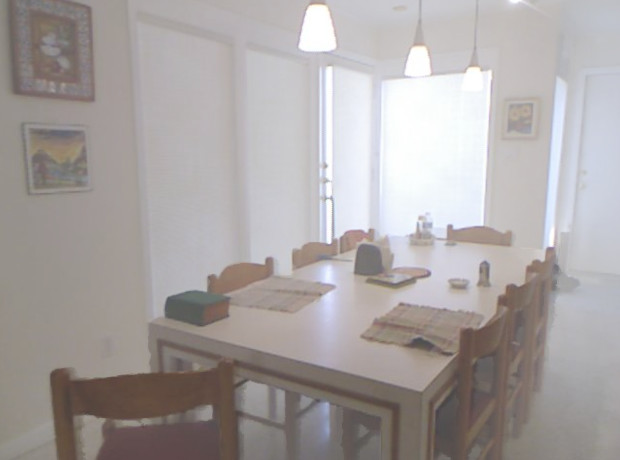}
            \centerline{-}
            \hspace{.05in}

            \includegraphics[width=0.95in]{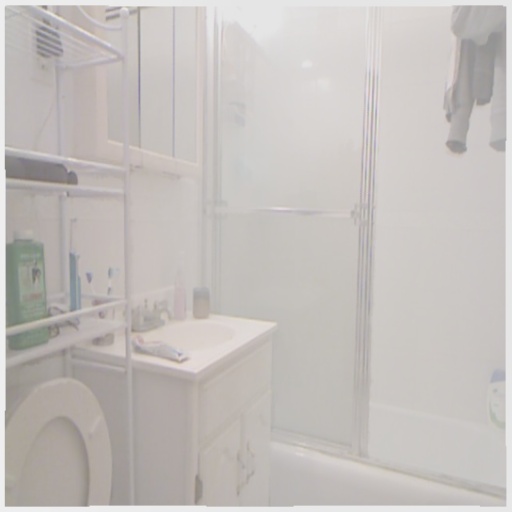}
            \centerline{-}
            \hspace{.05in}

            \includegraphics[width=0.95in]{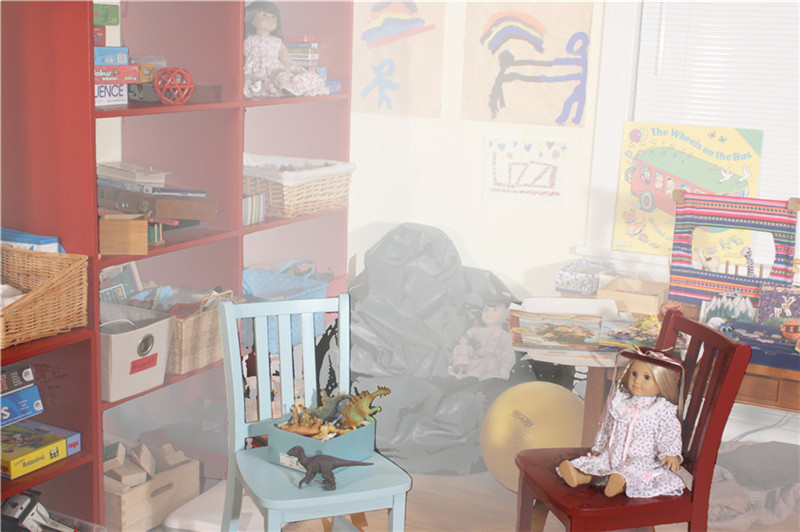}
            \centerline{-}
            \hspace{.05in}

            \includegraphics[width=0.95in]{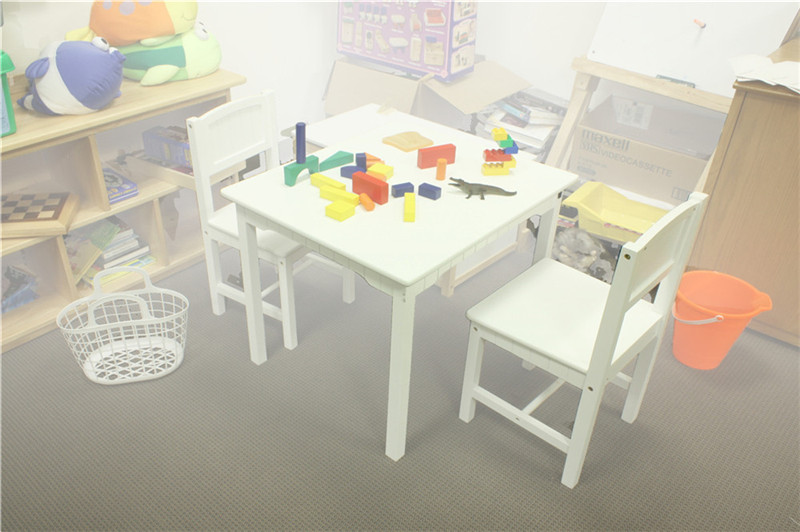}
            \centerline{-}
            \hspace{.05in}

            \includegraphics[width=0.95in]{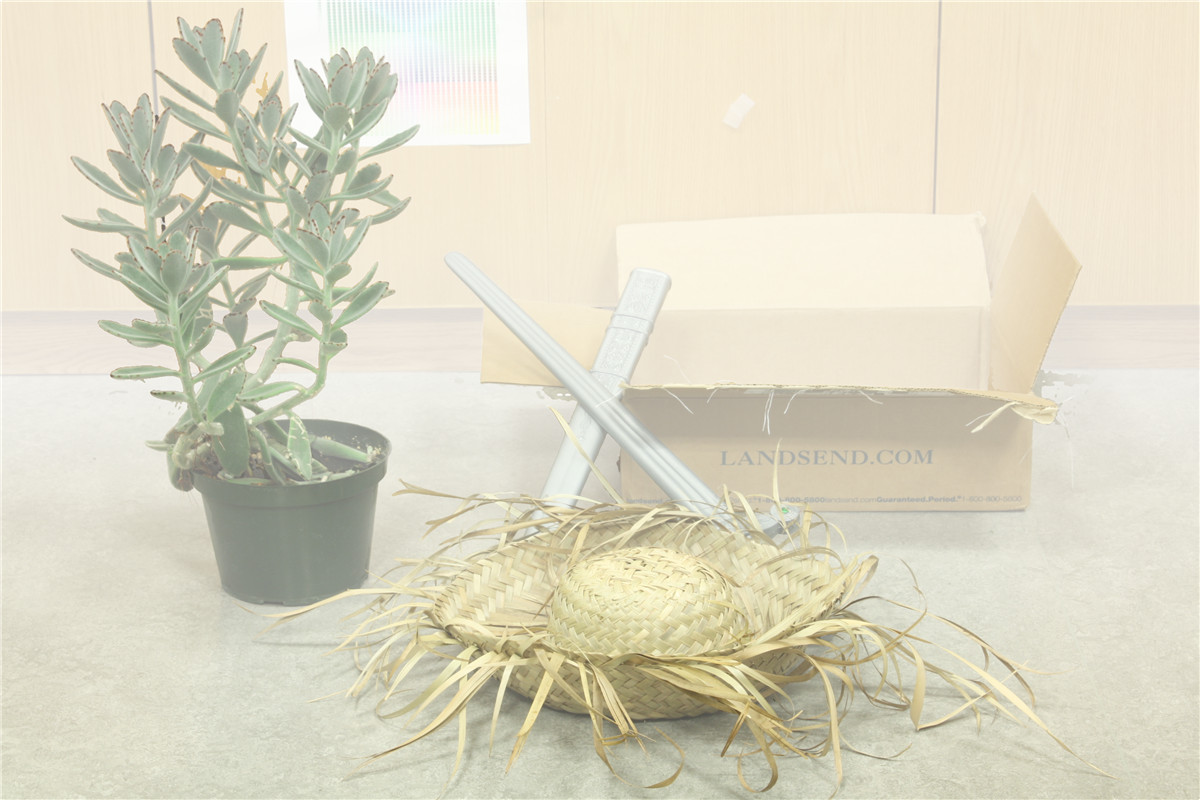}
            \centerline{-}
            \hspace{.05in}

        \end{minipage}%
}%
    \subfigure[DCP \cite{he2010single}]{
        \begin{minipage}[t]{0.135\linewidth}
            \centering%
            \includegraphics[width=0.95in]{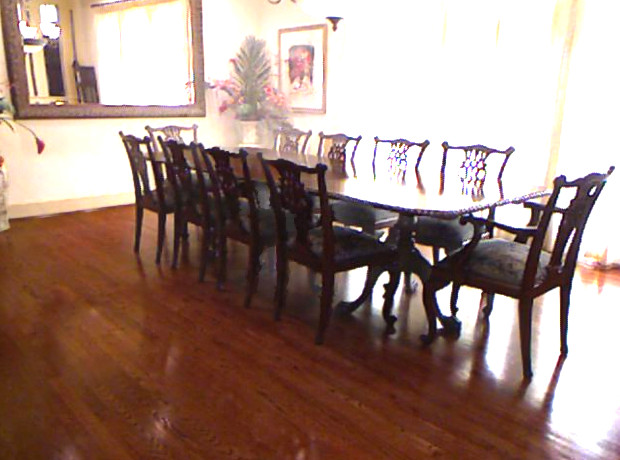}
            \centerline{14.62}
            \hspace{.05in}

            \includegraphics[width=0.95in]{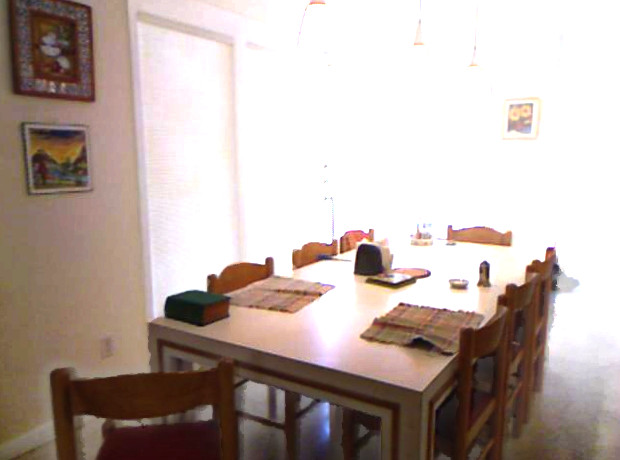}
            \centerline{12.39}
            \hspace{.05in}

            \includegraphics[width=0.95in]{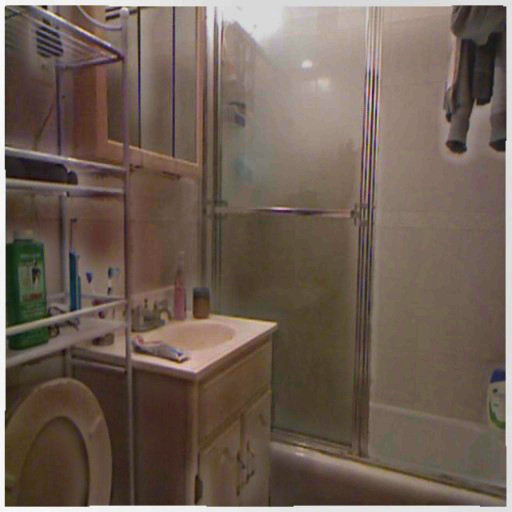}
            \centerline{16.79}
            \hspace{.05in}

            \includegraphics[width=0.95in]{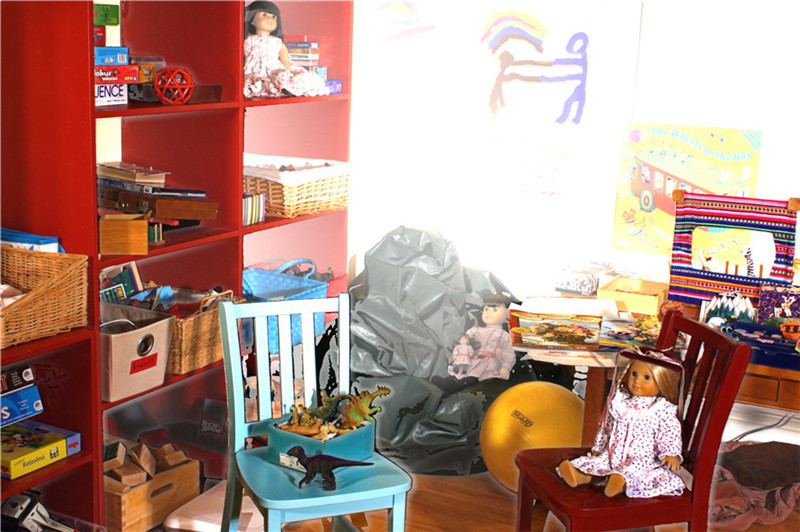}
            \centerline{13.35}
            \hspace{.05in}

            \includegraphics[width=0.95in]{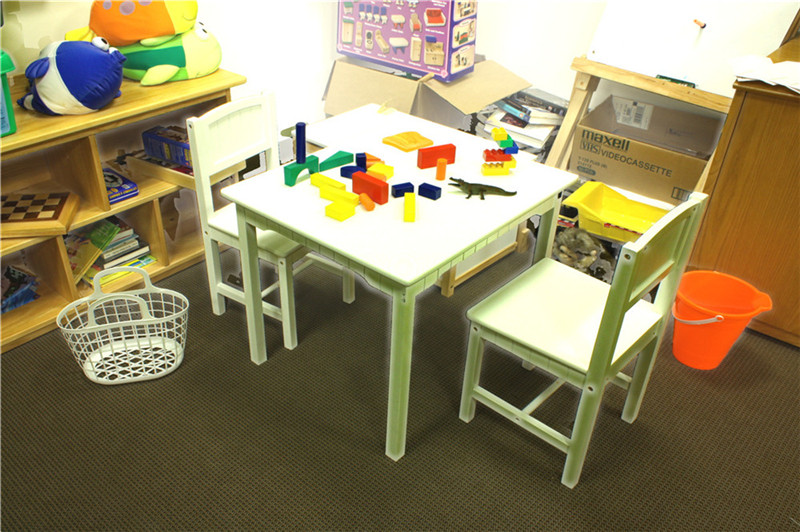}
            \centerline{16.70}
            \hspace{.05in}

            \includegraphics[width=0.95in]{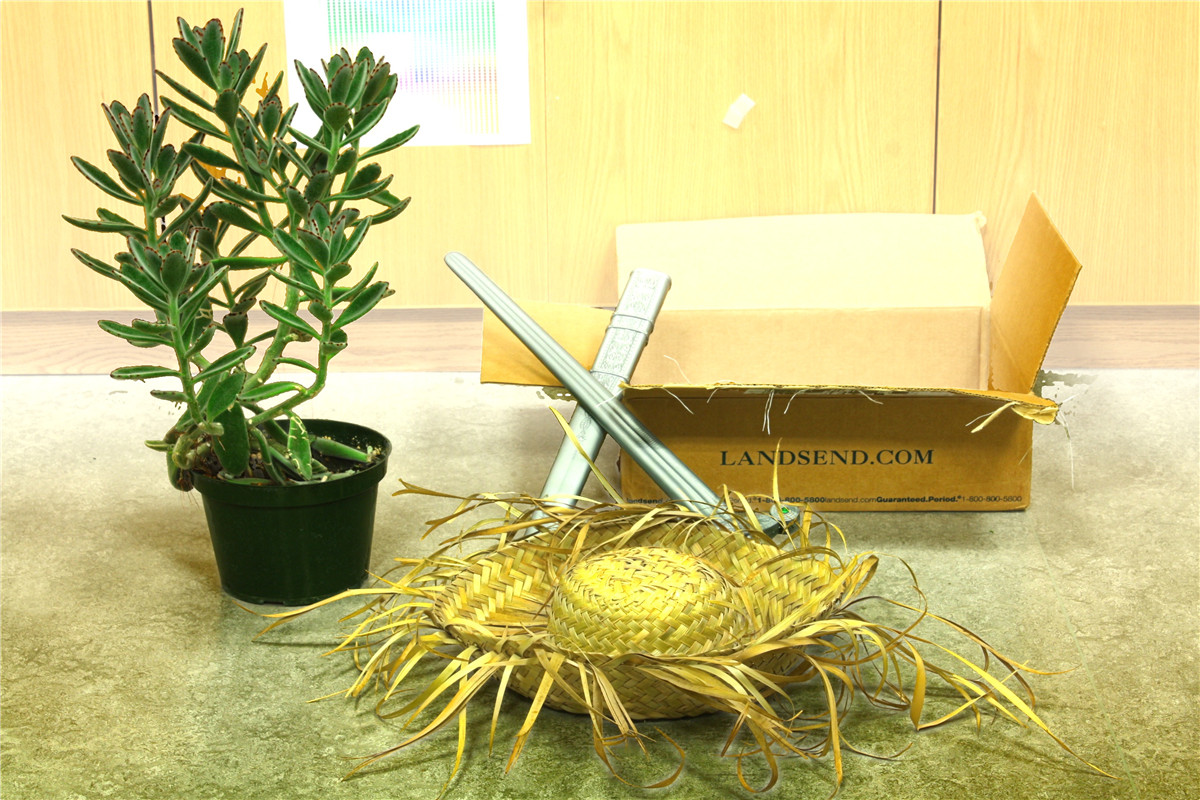}
            \centerline{14.64}
            \hspace{.05in}

        \end{minipage}%
}%
    \subfigure[AOD-Net \cite{li2017aod}]{
        \begin{minipage}[t]{0.135\linewidth}
            \centering%
            \includegraphics[width=0.95in]{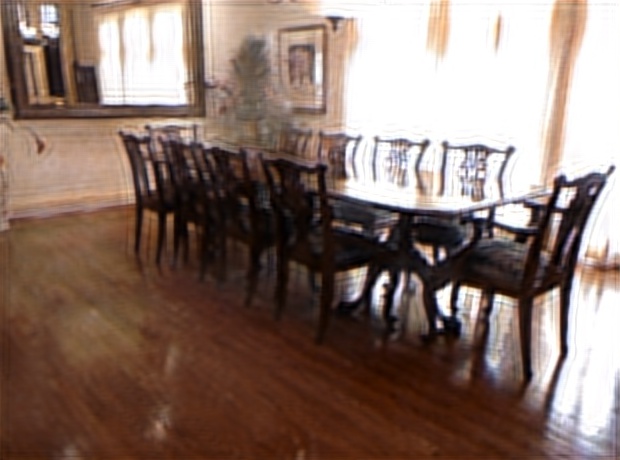}
            \centerline{20.20}
            \hspace{.05in}

            \includegraphics[width=0.95in]{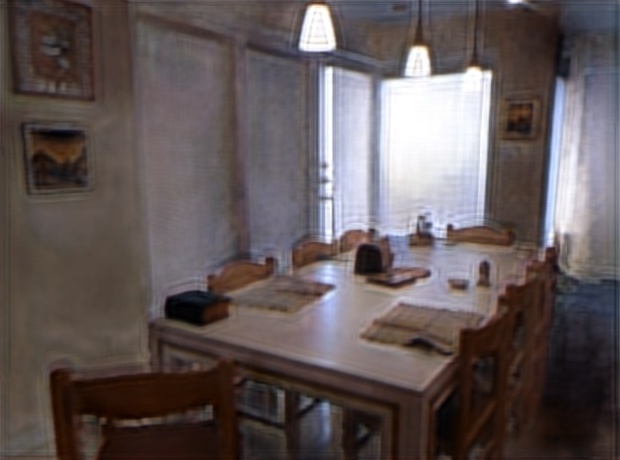}
            \centerline{19.90}
            \hspace{.05in}

            \includegraphics[width=0.95in]{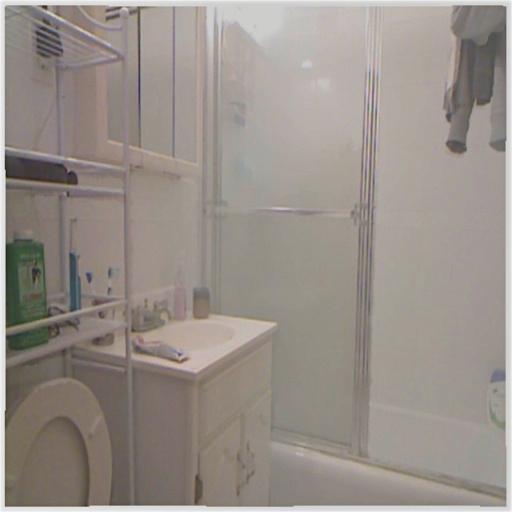}
            \centerline{20.84}
            \hspace{.05in}

            \includegraphics[width=0.95in]{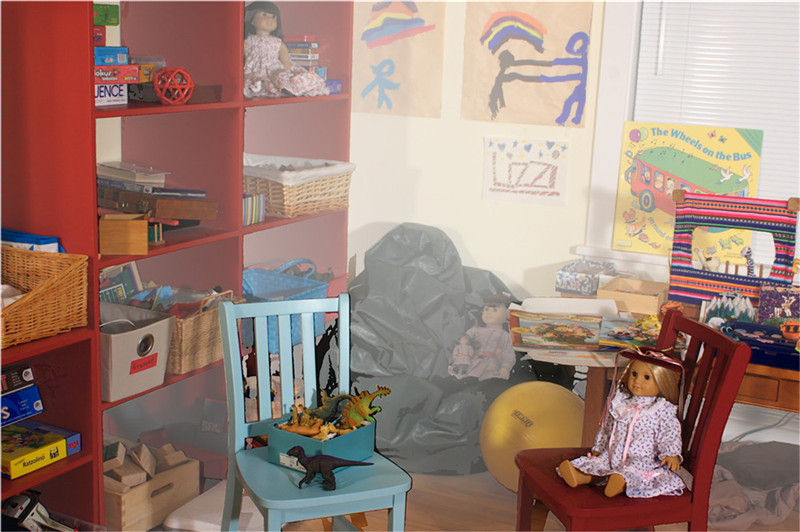}
            \centerline{15.45}
            \hspace{.05in}

            \includegraphics[width=0.95in]{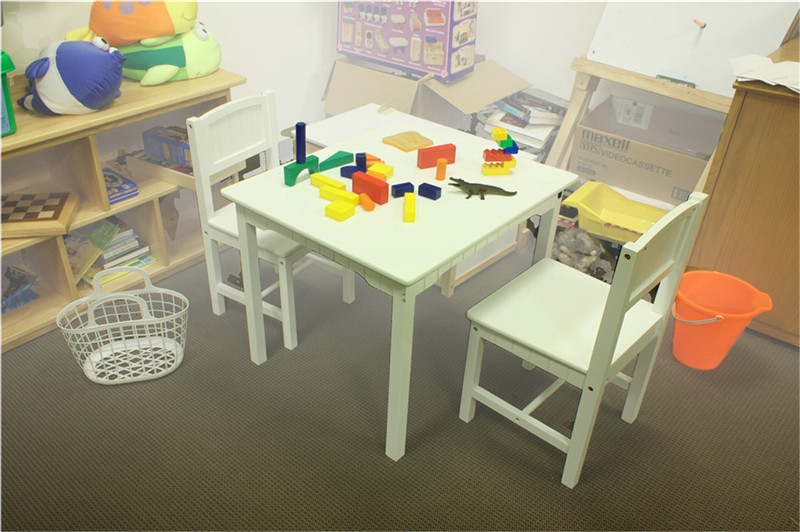}
            \centerline{16.76}
            \hspace{.05in}

            \includegraphics[width=0.95in]{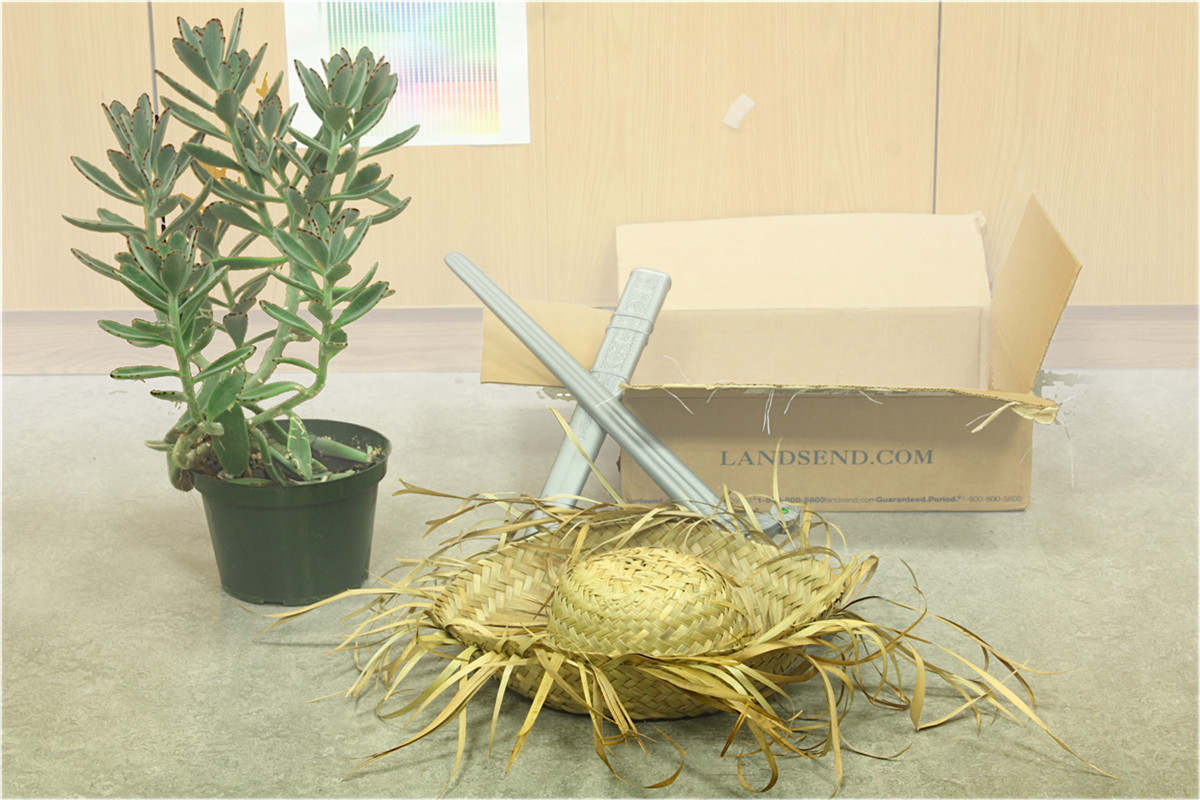}
            \centerline{15.42}
            \hspace{.05in}

        \end{minipage}%
}%
    \subfigure[GRID-Net \cite{liu2019griddehazenet}]{
        \begin{minipage}[t]{0.135\linewidth}
            \centering%
            \includegraphics[width=0.95in]{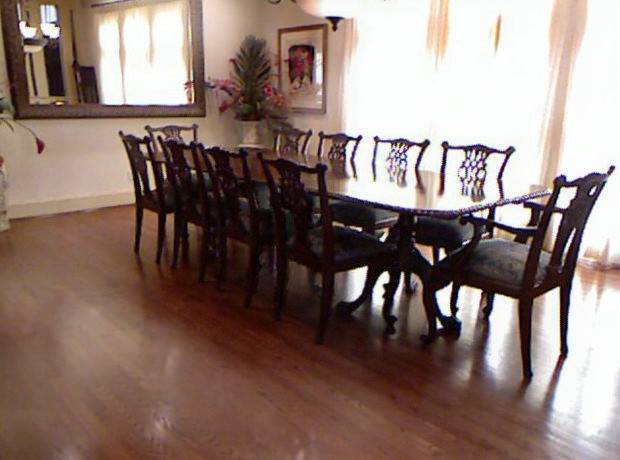}
            \centerline{27.21}
            \hspace{.05in}

            \includegraphics[width=0.95in]{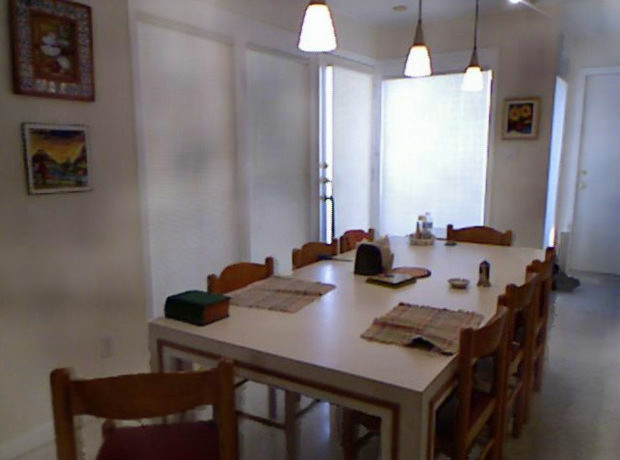}
            \centerline{26.08}
            \hspace{.05in}

            \includegraphics[width=0.95in]{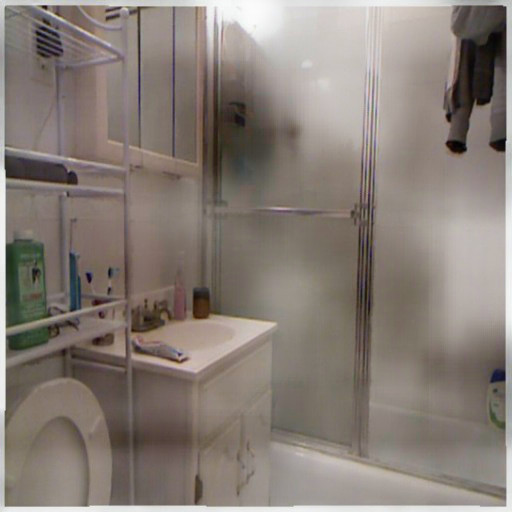}
            \centerline{20.68}
            \hspace{.05in}

            \includegraphics[width=0.95in]{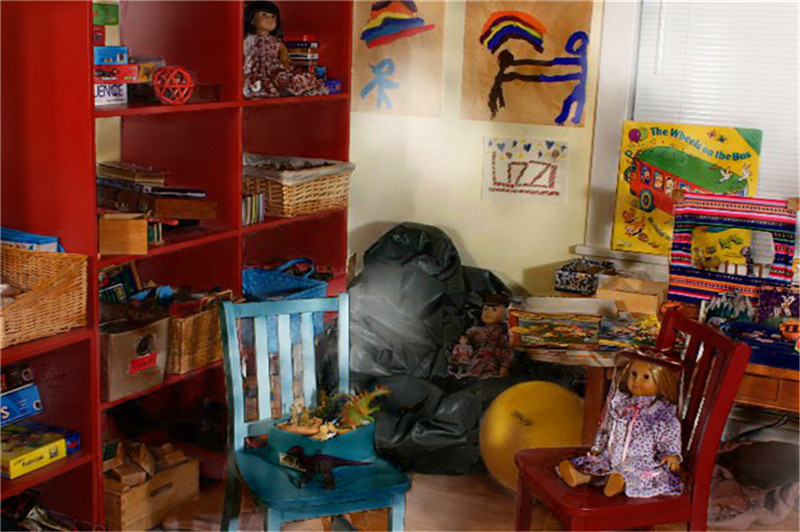}
            \centerline{16.37}
            \hspace{.05in}

            \includegraphics[width=0.95in]{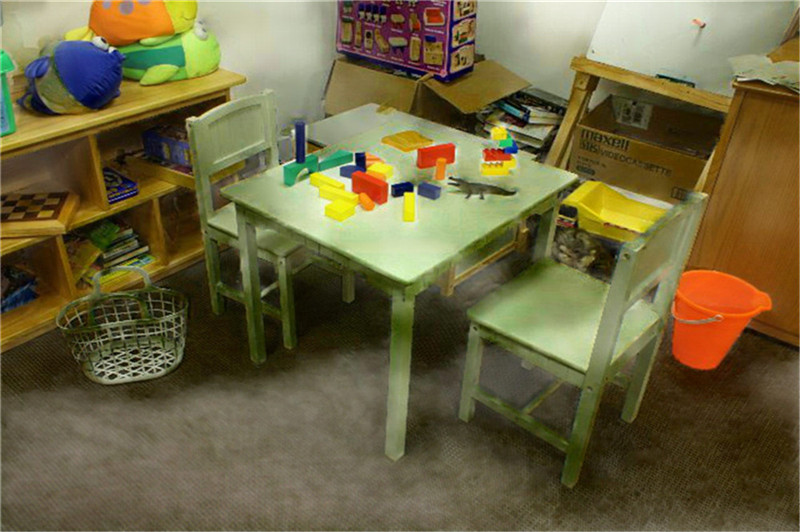}
            \centerline{15.97}
            \hspace{.05in}

            \includegraphics[width=0.95in]{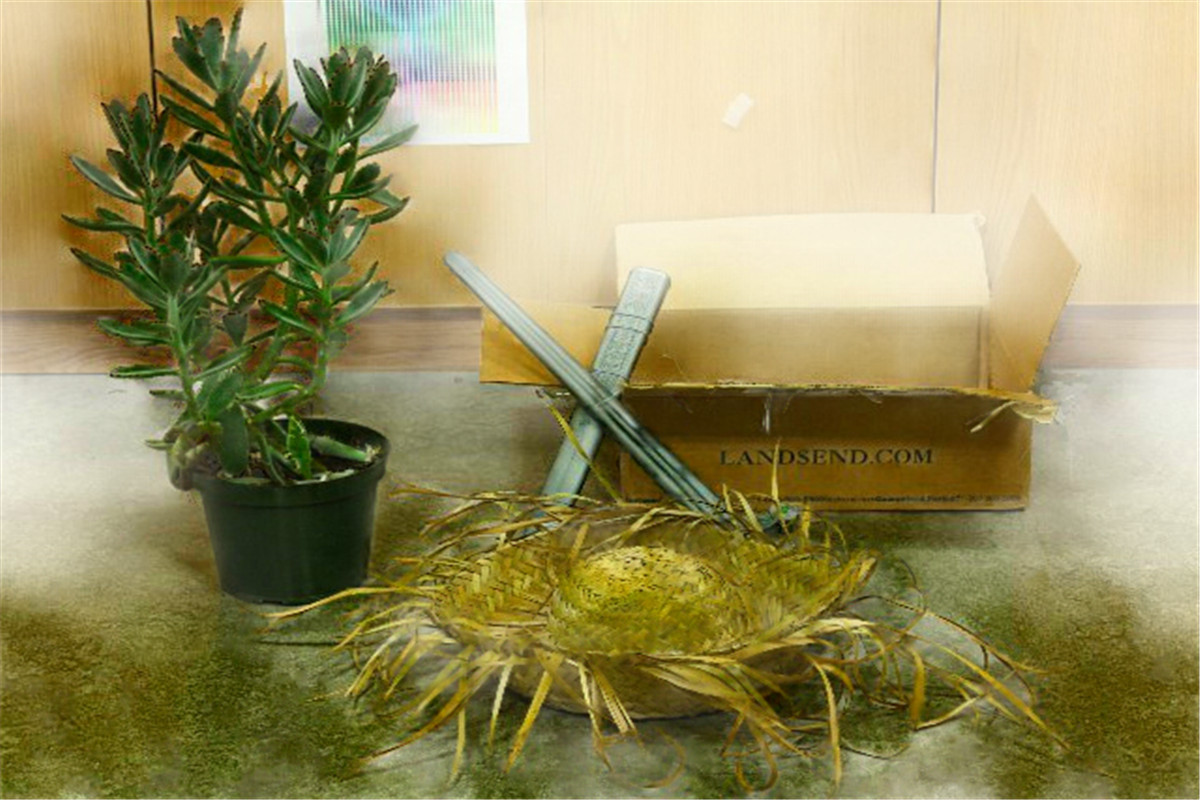}
            \centerline{11.28}
            \hspace{.05in}

        \end{minipage}%
}%
    \subfigure[FFA-Net \cite{qin2019ffa}]{
        \begin{minipage}[t]{0.135\linewidth}
            \centering%
            \includegraphics[width=0.95in]{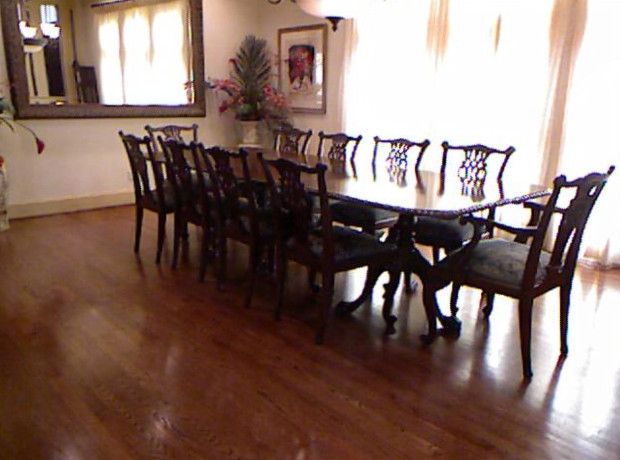}
            \centerline{34.59}
            \hspace{.05in}

            \includegraphics[width=0.95in]{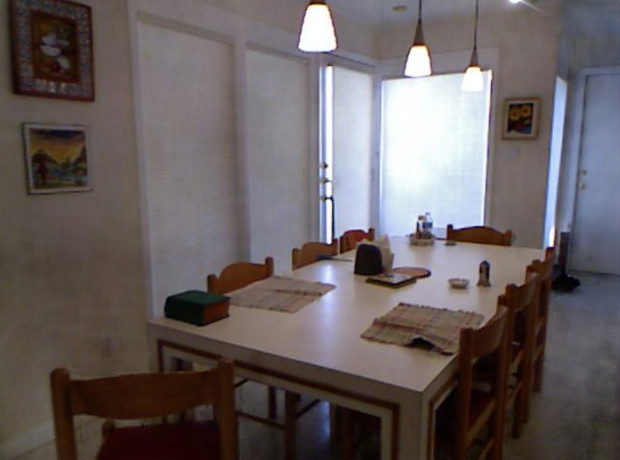}
            \centerline{33.67}
            \hspace{.05in}

            \includegraphics[width=0.95in]{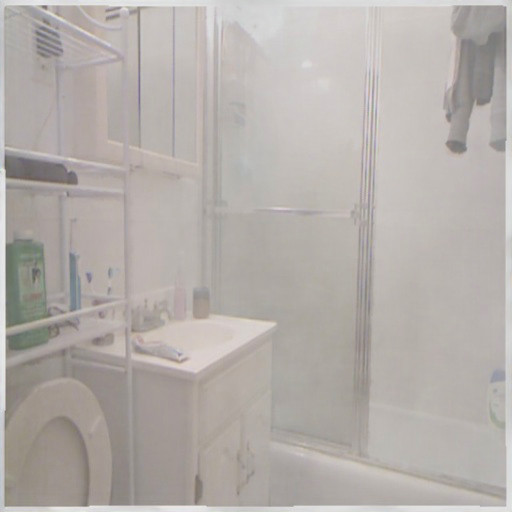}
            \centerline{18.87}
            \hspace{.05in}

            \includegraphics[width=0.95in]{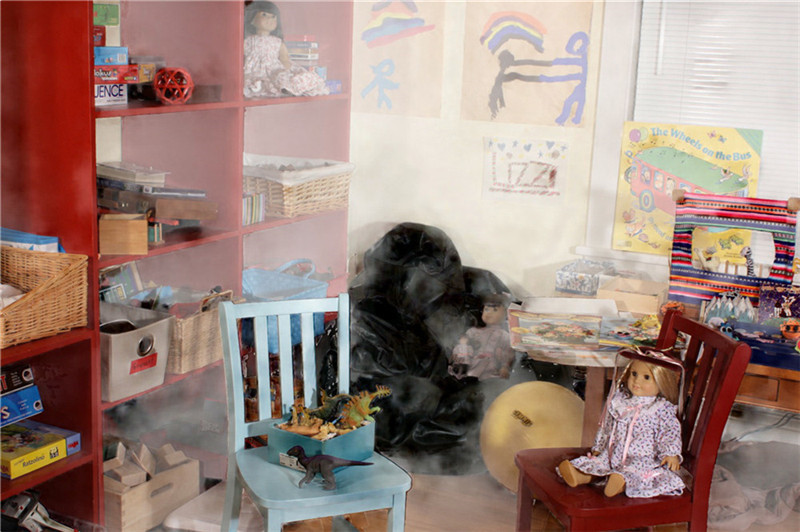}
            \centerline{14.50}
            \hspace{.05in}

            \includegraphics[width=0.95in]{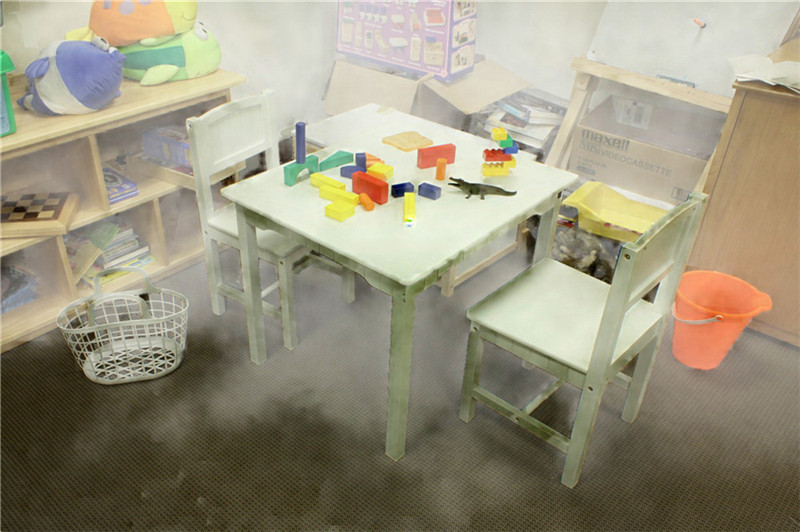}
            \centerline{14.23}
            \hspace{.05in}

            \includegraphics[width=0.95in]{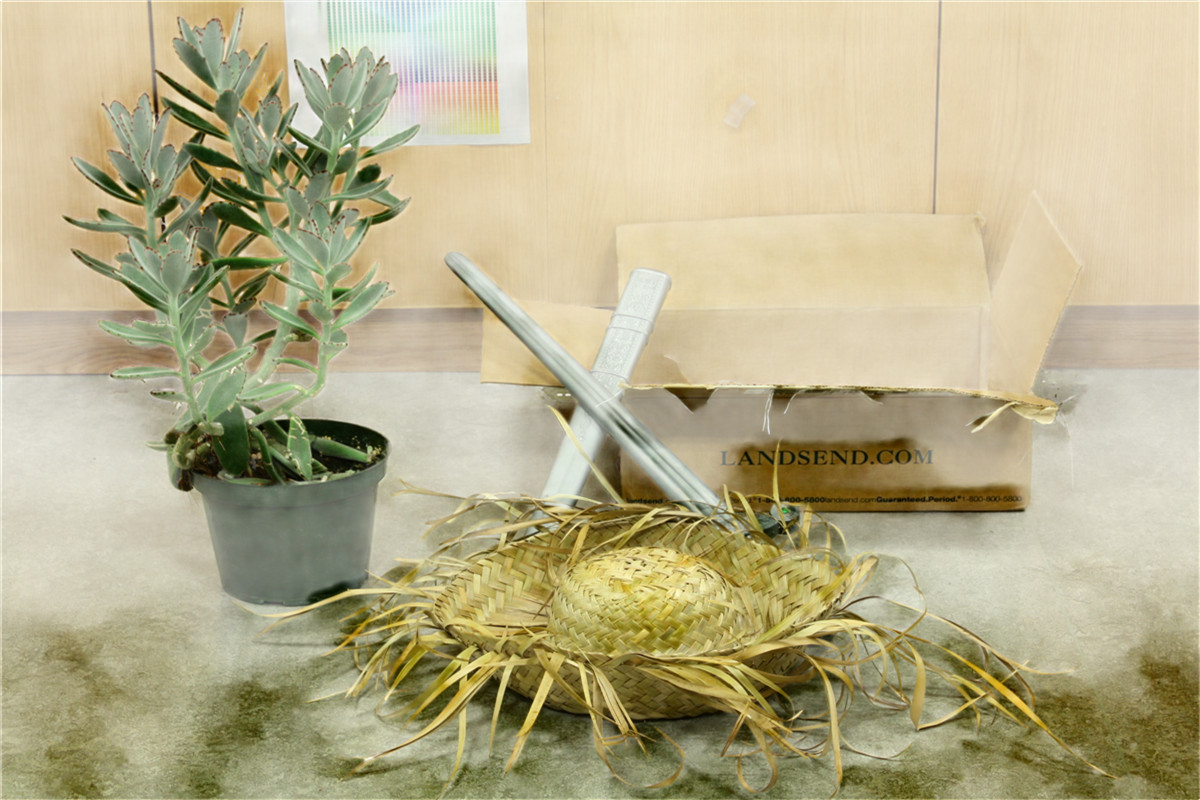}
            \centerline{13.27}
            \hspace{.05in}

        \end{minipage}%
}%
    \subfigure[OURS]{
        \begin{minipage}[t]{0.135\linewidth}
            \centering%
            \includegraphics[width=0.95in]{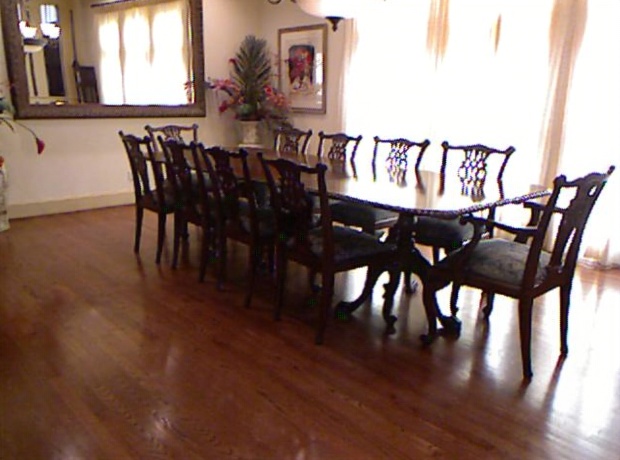}
            \centerline{34.61}
            \hspace{.05in}

            \includegraphics[width=0.95in]{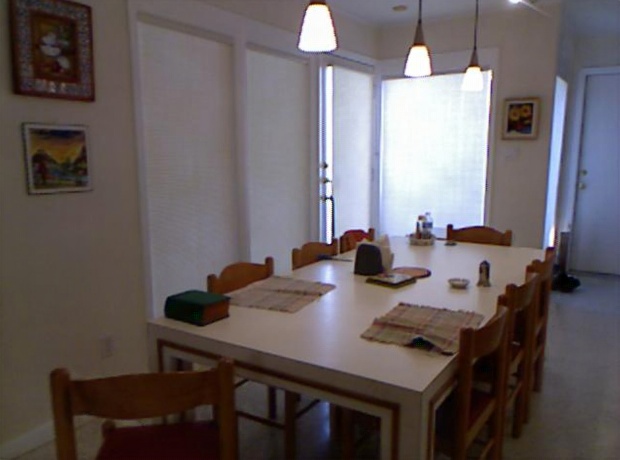}
            \centerline{35.34}
            \hspace{.05in}

            \includegraphics[width=0.95in]{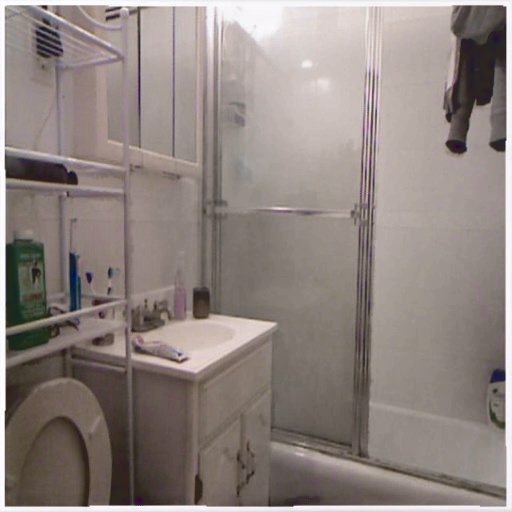}
            \centerline{25.43}
            \hspace{.05in}

            \includegraphics[width=0.95in]{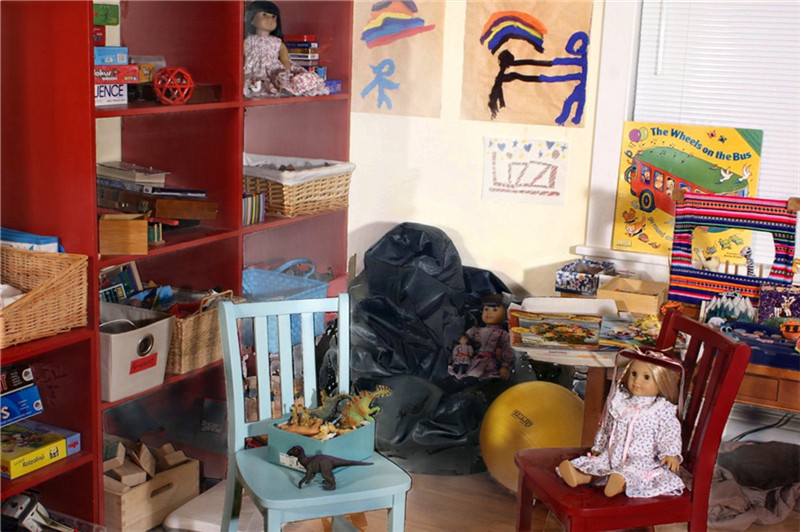}
            \centerline{19.42}
            \hspace{.05in}

            \includegraphics[width=0.95in]{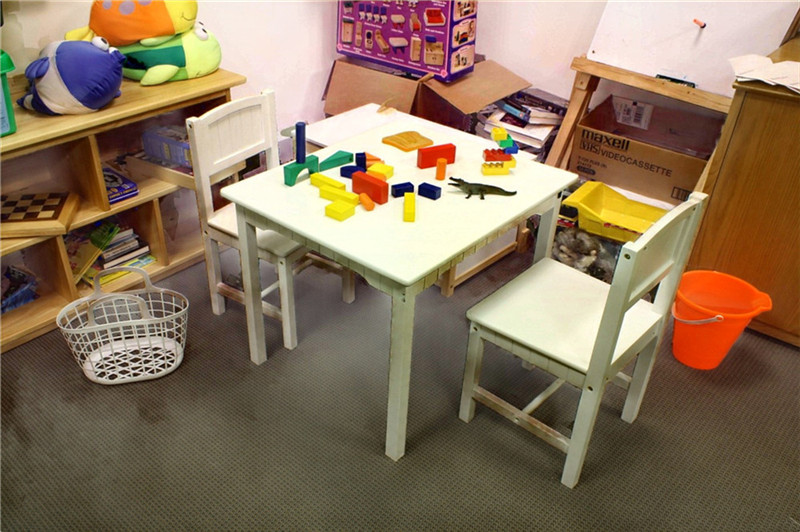}
            \centerline{19.86}
            \hspace{.05in}

            \includegraphics[width=0.95in]{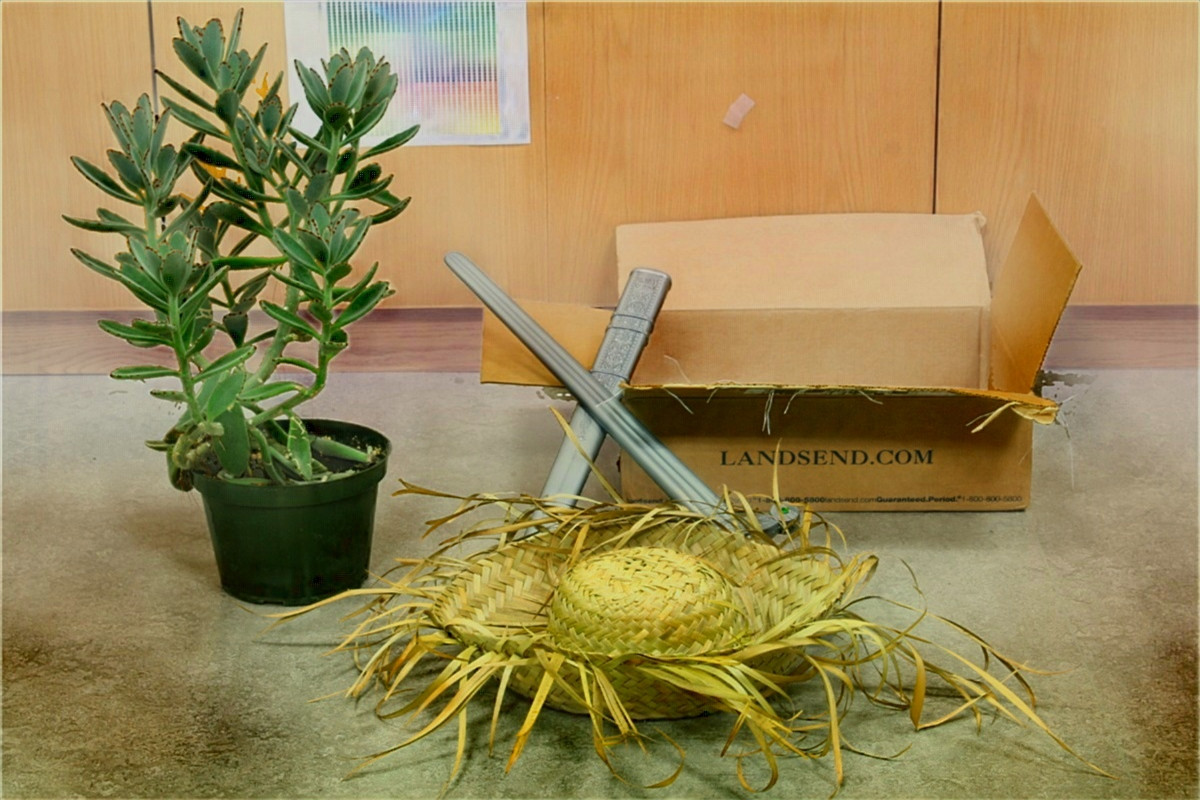}
            \centerline{17.64}
            \hspace{.05in}

        \end{minipage}%
}%
    \subfigure[GT]{
        \begin{minipage}[t]{0.135\linewidth}
            \centering%
            \includegraphics[width=0.95in]{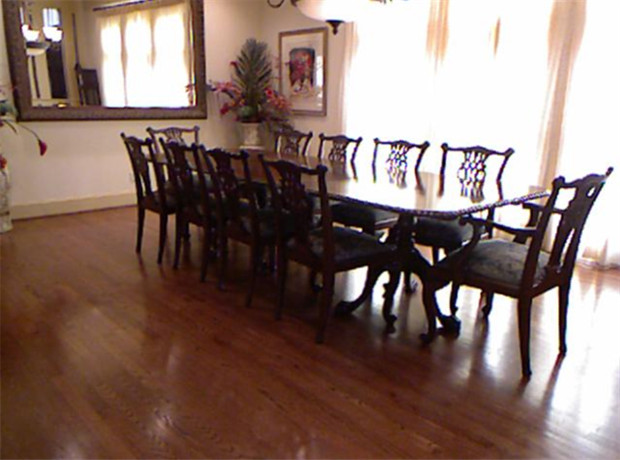}
            \centerline{$\infty$}
            \hspace{.05in}

            \includegraphics[width=0.95in]{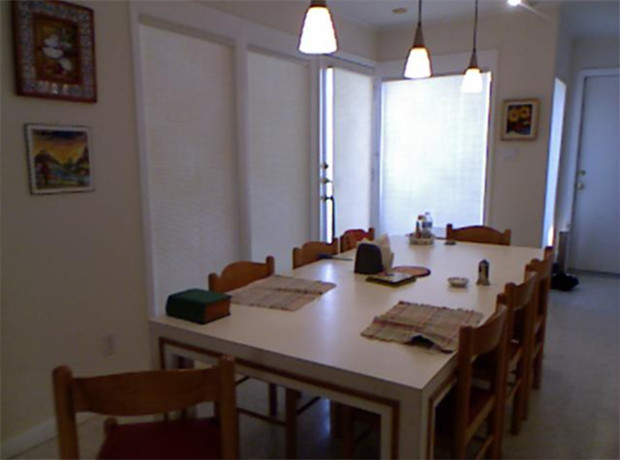}
            \centerline{$\infty$}
            \hspace{.05in}

            \includegraphics[width=0.95in]{Aclean.jpg}
            \centerline{$\infty$}
            \hspace{.05in}

            \includegraphics[width=0.95in]{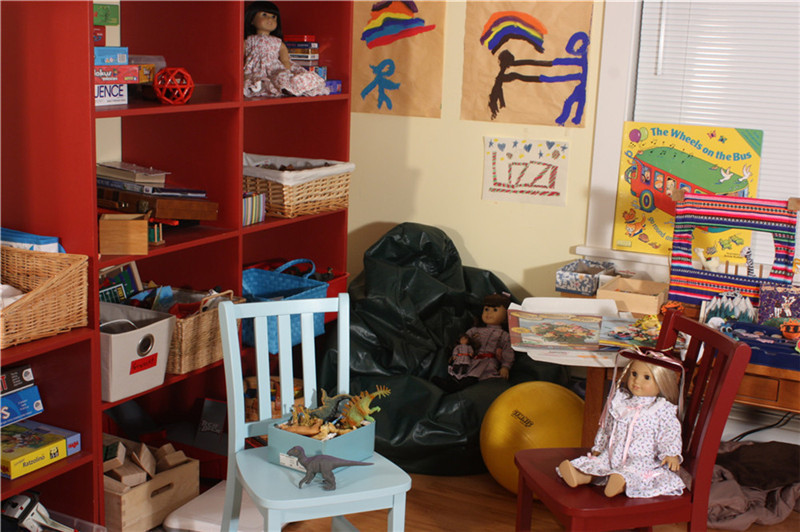}
            \centerline{$\infty$}
            \hspace{.05in}

            \includegraphics[width=0.95in]{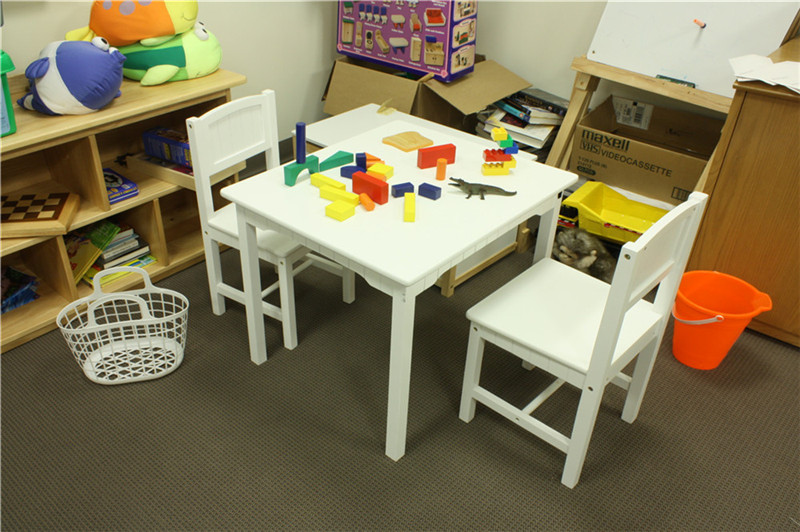}
            \centerline{$\infty$}
            \hspace{.05in}

            \includegraphics[width=0.95in]{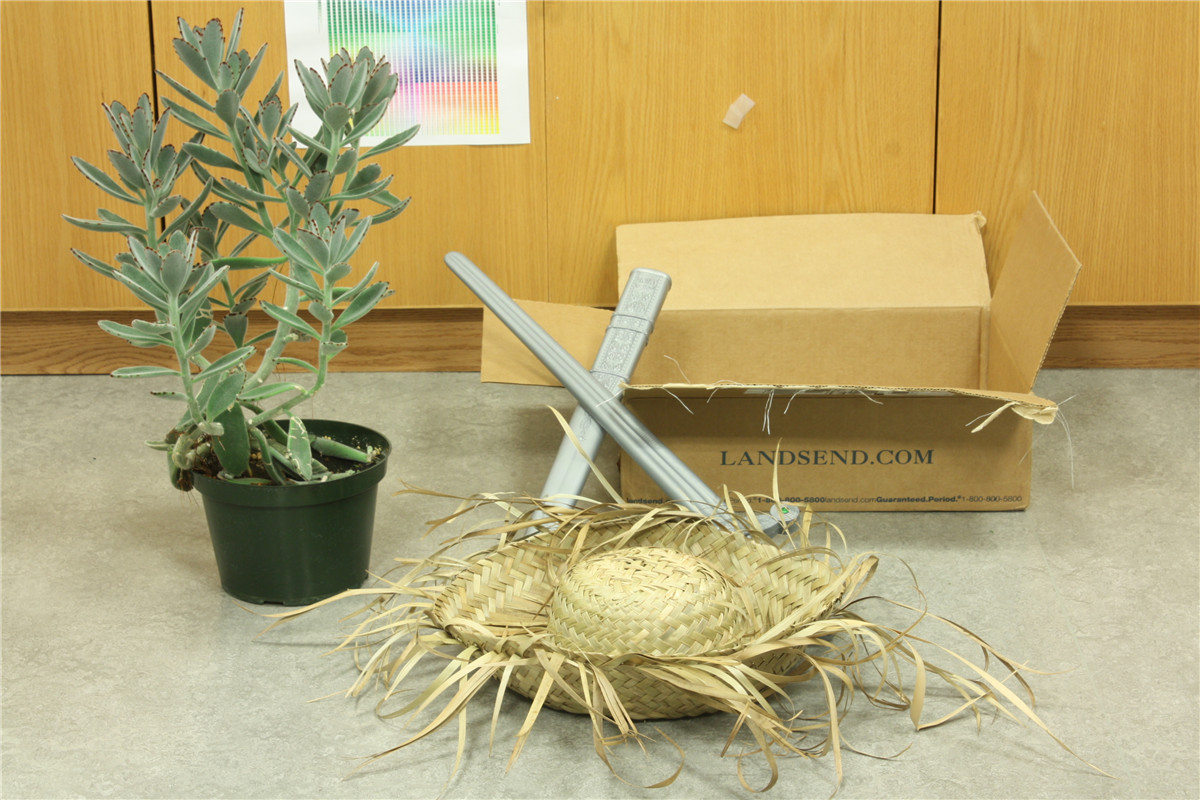}
            \centerline{$\infty$}
            \hspace{.05in}

        \end{minipage}
}%

\centering
\caption{Qualitative comparisons with different state-of-the-art dehazing methods for indoor synthesis hazy images. The top two rows are from SOTS, the third row is from TestA dataset and the bottom three rows are from MiddleBury dehazing dataset. The numbers below image are PSNR (dB) value of each image.}
\label{synthesis}

\end{figure*}

\subsection{Residual Channel Attention block}
The fused feature $x$ is inevitably mixed up with artifacts generated during the process of CNN, which would degrade the final performance significantly. We note that the existed disturbance would reduce the information of each channel in feature map has and the distribution of artifacts in each channel is uneven. The channel attention mechanism provides insight into this problem. Inspired by \cite{hu2018squeeze}, we propose a modified residual attention block (RCA-block) to mitigate this. RCA-block generates an attention map to reweigh feature map of each channel, which treats channel unequally thus provides extra flexibility in suppressing background disturbance.

Specially, as illustrated in Fig. \ref{RCA}, RCA-block firstly generates attention map $\Psi$ from the fused features $x$:
\begin{align}
   \Psi=\textrm{sigmoid}(\textrm{MLP}(\textrm{GAP}(\textrm{conv}(x)),\Psi \in \mathbb R^{c\times1\times1}
\end{align}
where $\textrm{conv}$ is $1\times1$ convolutional layers. Then Global Average Pooling (GAP) $(\mathbb{R}^{c\times h\times w}\rightarrow \mathbb{R}^{c\times1\times1})$ is applied to convert the global spatial information into a channel descriptor. To obtain the attention map, channel descriptor passes through multi-layer perceptron (MLP), which consists of two fully connected layers, and sigmoid activation function.

Then the input $x$ element-wise multiplies $\circ$ with the attention map $\Psi$ then adds input $x$ to get the refined feature $x_r$:
\begin{align}
    x_r = \Psi \circ x + x
\end{align}
where each weight in attention map $\Psi$ where $\Psi \in \mathbb R^{c\times1\times1}$ reflects the information in the corresponding channel. A higher value of weight indicates more information channel has. Therefore, multiplying feature with attention map $\Psi$ can suppress background noise effectively.

Considering the back propagation, operation of adding allows network firstly rely on the cue of the local feature then gradually learn parameters of attention map $\Psi$ in a global view.

\section{Experiments}

\begin{figure*}[]\setcounter{subfigure}{0}
    \centering%
        \subfigure[INPUT]{
        \begin{minipage}[t]{0.16\linewidth}
            \centering
            \includegraphics[width=1.1in]{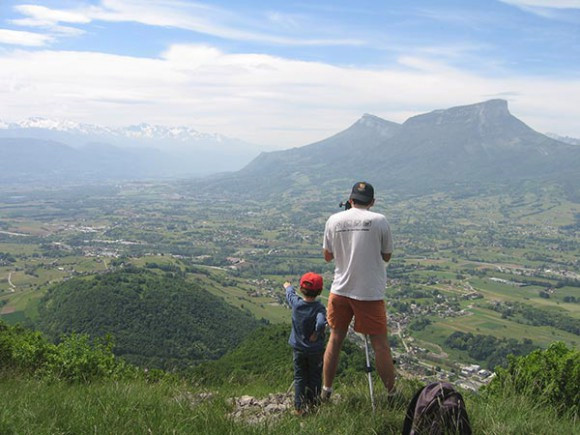}
            \hspace{.05in}

            \includegraphics[width=1.1in]{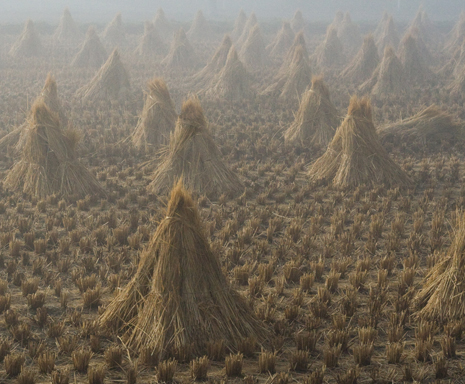}
            \hspace{.05in}

            \includegraphics[width=1.1in]{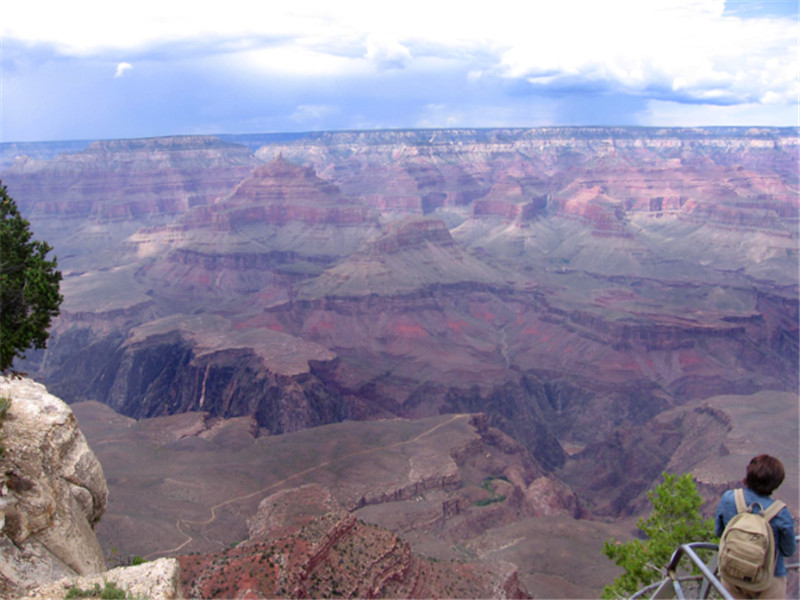}
            \hspace{.05in}

        \end{minipage}%
}%
    \subfigure[DCP \cite{he2010single}]{
        \begin{minipage}[t]{0.16\linewidth}
            \centering%
            \includegraphics[width=1.1in]{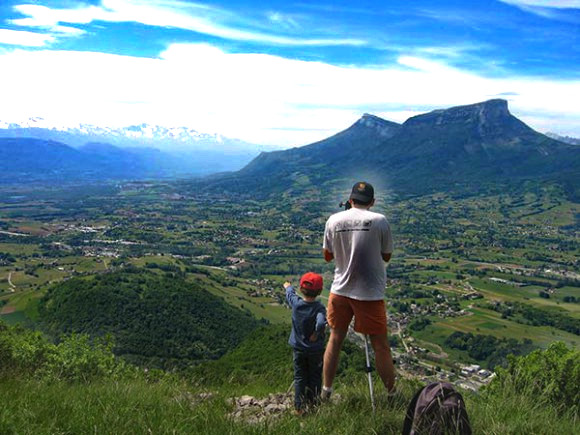}
            \hspace{.05in}

            \includegraphics[width=1.1in]{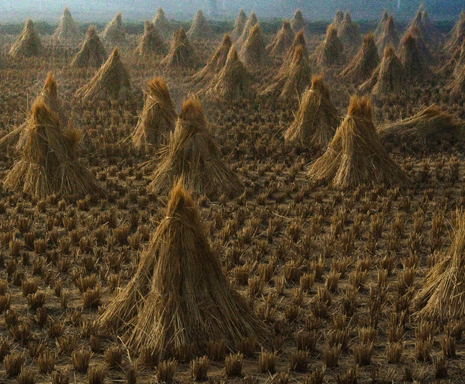}
            \hspace{.05in}

            \includegraphics[width=1.1in]{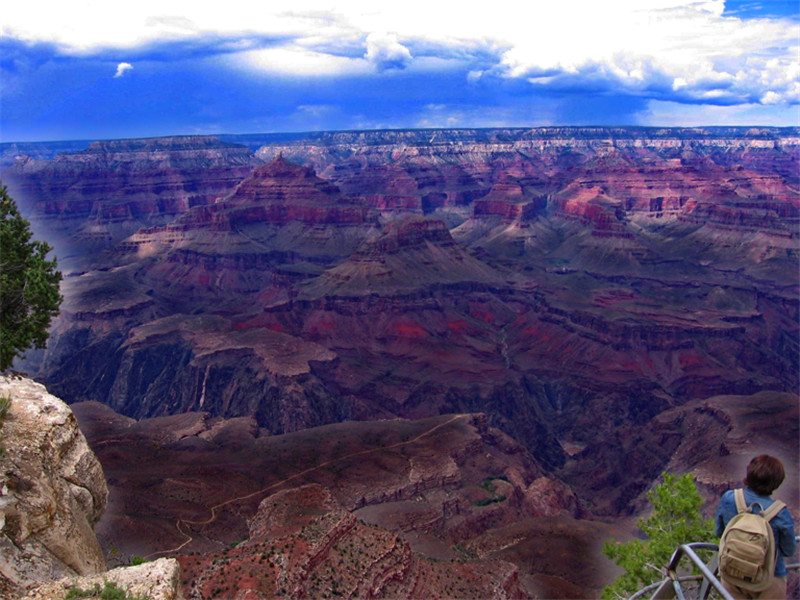}
            \hspace{.05in}

        \end{minipage}%
}%
    \subfigure[GRID-Net \cite{liu2019griddehazenet}]{
        \begin{minipage}[t]{0.16\linewidth}
            \centering%
            \includegraphics[width=1.1in]{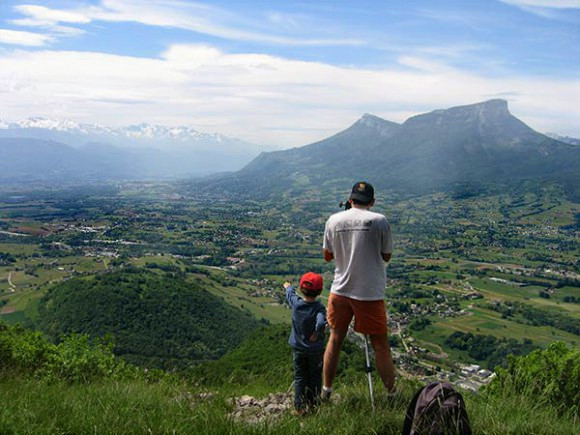}
            \hspace{.05in}

            \includegraphics[width=1.1in]{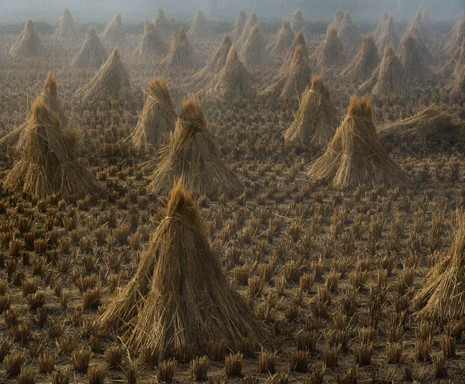}
            \hspace{.05in}

            \includegraphics[width=1.1in]{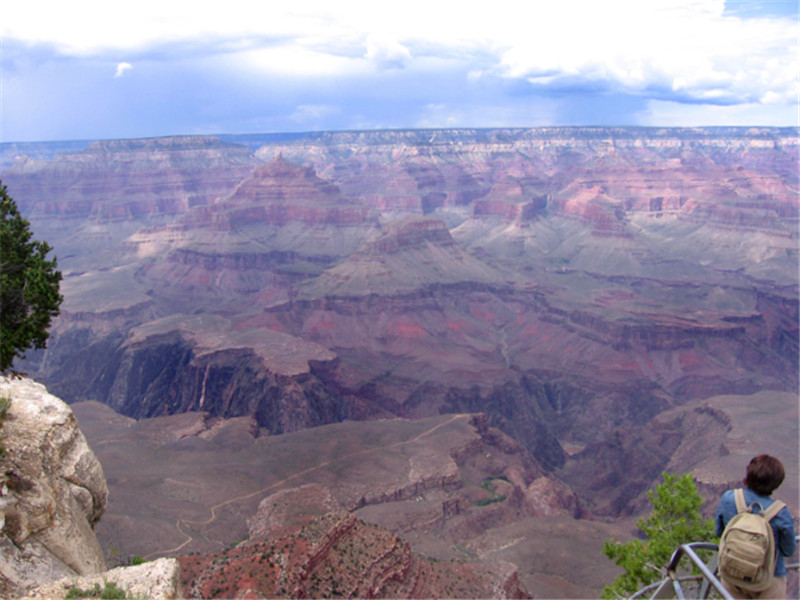}
            \hspace{.05in}

        \end{minipage}%
}%
    \subfigure[FFA-Net \cite{qin2019ffa}]{
        \begin{minipage}[t]{0.16\linewidth}
            \centering%
            \includegraphics[width=1.1in]{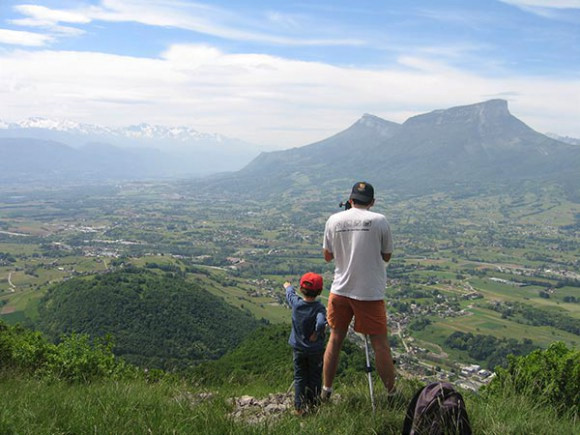}
            \hspace{.05in}

            \includegraphics[width=1.1in]{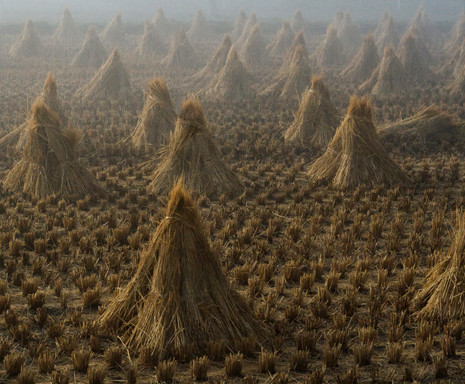}
            \hspace{.05in}

            \includegraphics[width=1.1in]{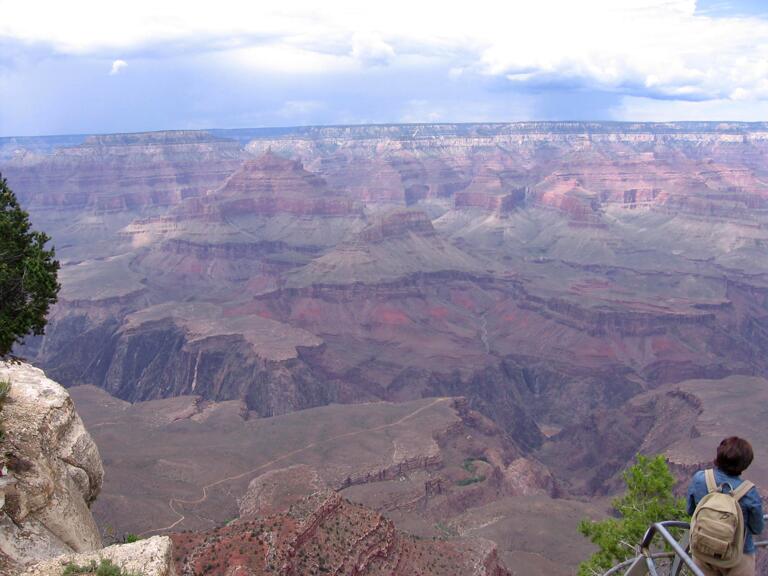}
            \hspace{.05in}

        \end{minipage}%
}%
    \subfigure[FD-GAN \cite{dong2020fd}]{
        \begin{minipage}[t]{0.16\linewidth}
            \centering%
            \includegraphics[width=1.1in]{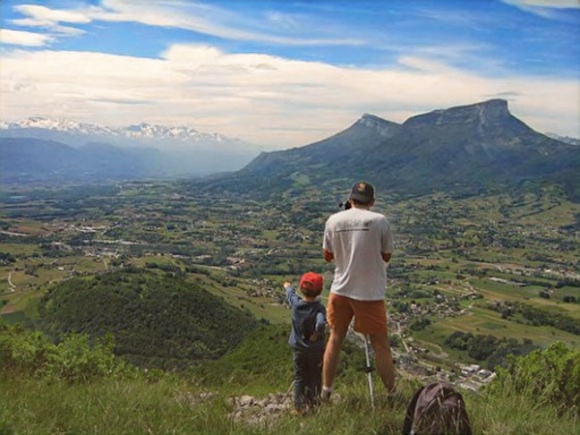}
            \hspace{.05in}

            \includegraphics[width=1.1in]{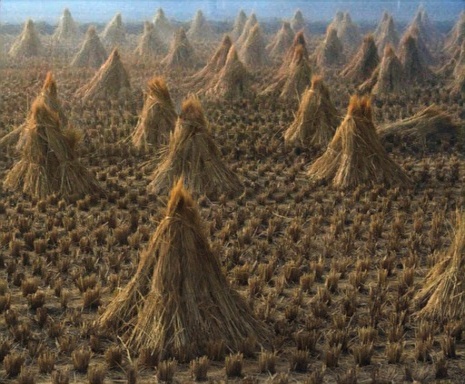}
            \hspace{.05in}

            \includegraphics[width=1.1in]{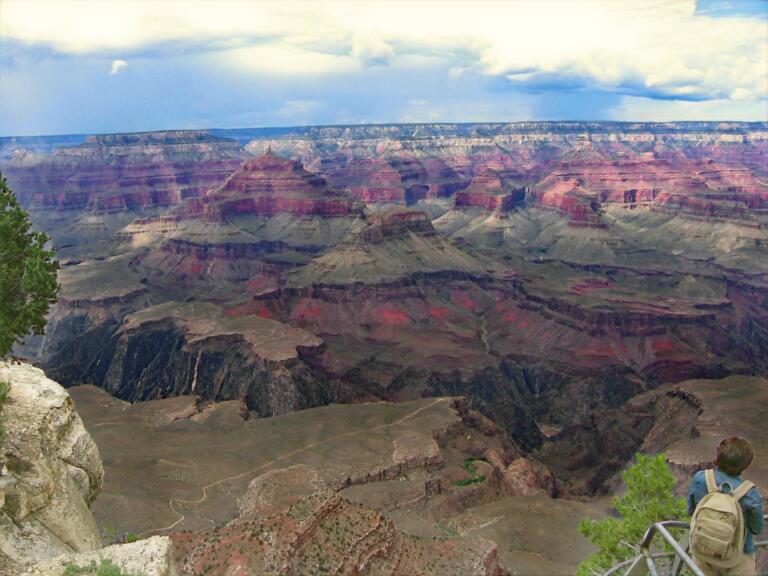}
            \hspace{.05in}

        \end{minipage}%
}%
    \subfigure[OURS]{
        \begin{minipage}[t]{0.16\linewidth}
            \centering%
            \includegraphics[width=1.1in]{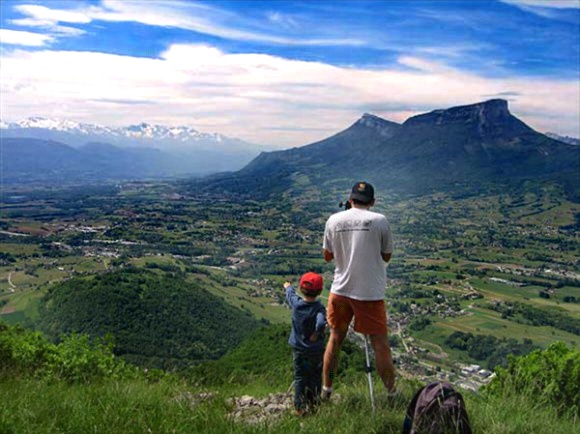}
            \hspace{.05in}

            \includegraphics[width=1.1in]{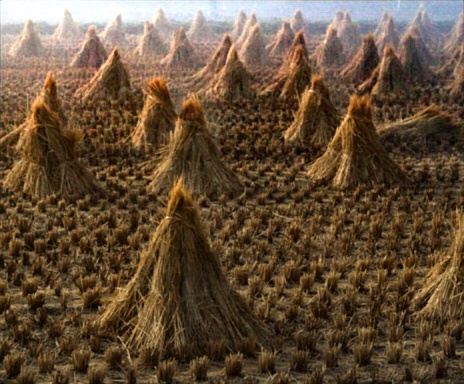}
            \hspace{.05in}

            \includegraphics[width=1.1in]{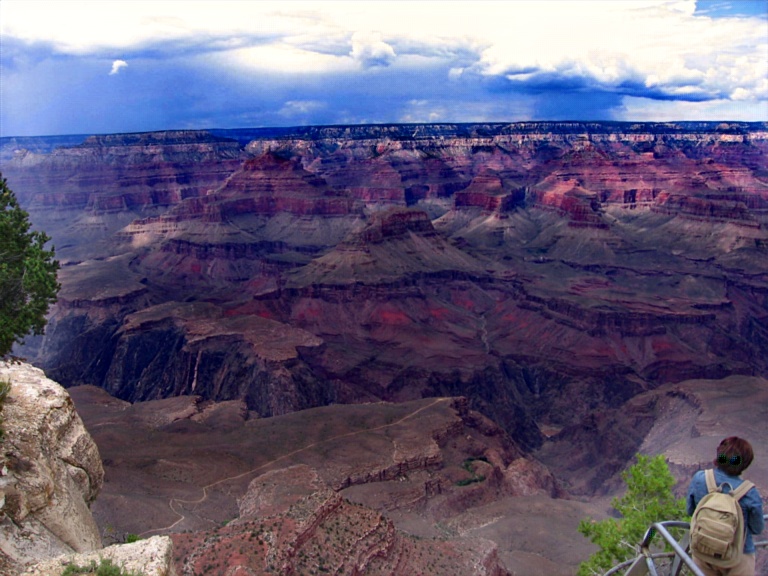}
            \hspace{.05in}

        \end{minipage}%
}%
\centering
\caption{Qualitative comparisons with different dehazing state-of-the-art methods for real hazy images.}
\label{real}
\end{figure*}

\begin{figure}[t]
    \centering
    \includegraphics[scale=0.19]{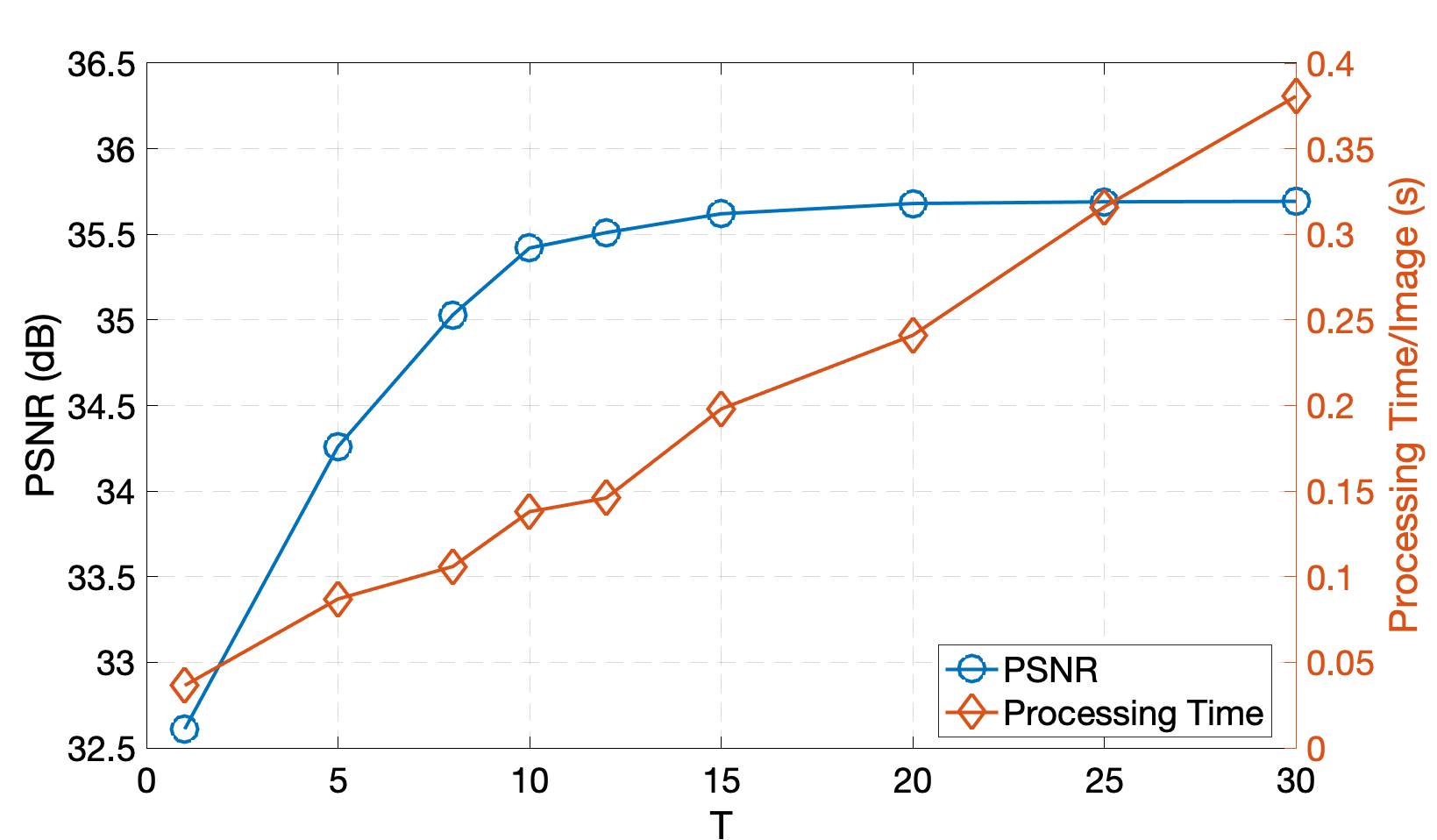}
    \caption{Performance and processing time curve for MI-Net with different recursion number $T$ for IM-blocks. The processing time is tested with NVIDIA GTX 1060 GPU.}
    \label{iteration}
\end{figure}
\subsection{Datasets and Metrics}
We use the benchmark synthetic dataset RESIDE \cite{li2018benchmarking}, TestA \cite{zhang2018densely} and Middlebury\cite{Westling2014High} for the evaluation of our method. Indoor Training Set (ITS) and Synthetic Objective Testing set (SOTS) from RESIDE are adopted in our experiment. ITS which contains 13,500 synthetic indoor hazy images is applied as the training set. SOTS consists of 500 indoor images and 200 outdoor images with light, medium and high level of haze. TestA is a synthesis dataset introduced by DCPDN. Due to the low resolution and similar scene of images in the SOTS testing set and TestA, we adopt Middlebury synthetic dataset, a high-resolution stereo datasets with subpixel-accurate ground truth, as an assistant testing set. The high resolution of image in Middlebury can provide abundant processing details. Besides, we have done evaluation on real-world images to validate the performance of our proposed network. The dehazing performance on synthetic dataset is evaluated with Peak Signal-to-Noise Ratio (PSNR) and Structural Similarity (SSIM)\cite{wang2004image}. Due to the lack of groundtruth of the real-world images, real-world images are evaluated by visualization.

\begin{figure}[]\setcounter{subfigure}{0}
    \centering%
        \subfigure[Origin]{
        \begin{minipage}[t]{0.32\linewidth}
            \centering
            \includegraphics[width=1.05in]{REALHAZY1.jpg}
            \hspace{.05in}
        \end{minipage}%
}%
        \subfigure[IM-Net (OTS)]{
        \begin{minipage}[t]{0.32\linewidth}
            \centering
            \includegraphics[width=1.05in]{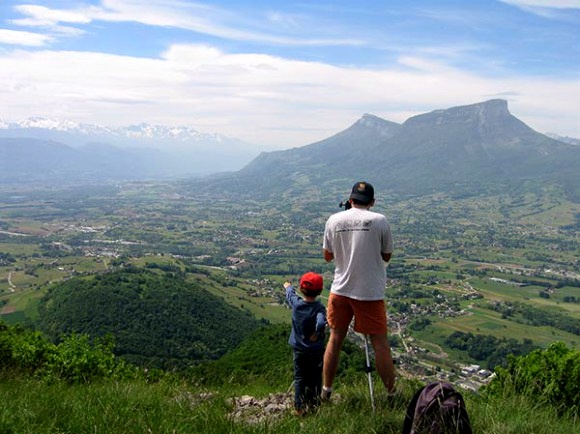}
            \hspace{.05in}
        \end{minipage}%
}%
        \subfigure[IM-Net (ITS)]{
        \begin{minipage}[t]{0.32\linewidth}
            \centering
            \includegraphics[width=1.05in]{OURSreal1.jpg}
            \hspace{.05in}
        \end{minipage}%
}%
\caption{The evaluation of our model trained on indoor training set (ITS) and outdoor training dataset (OTS) from RESIDE. For IM-Net (OTS), the haze in foreground is removed but not for the background.}
\label{out}
\end{figure}
\subsection{Implementation Details}
For training, we directly use RGB images as input instead of using image patches since image patches will lose structure information of original image. The parameter settings for our proposed MI-Net are as follows. For the encoding and decoding blocks shown in Fig.~\ref{MINET}, the convolutional layers have 3 filters of size $3\times 3$. While for the other blocks, each convolution layer has 64 filters of size $3\times 3$. Besides, The dilated rate of MLE blocks are $\times 1$, $\times 2$, and $\times 5$ respectively. The zero padding is adopted to fix the size of feature maps. We use Adam optimizer with $\beta_1=0.99$, $\beta_2=0.999$ to train our MI-Net for $350,000$ iterations. The learning rate is initially set to $1e^{-3}$, and then decays by 0.1 every $20,000$ iterations. Two NVIDIA TITAN RTX are used for the training phase and one NVIDIA GTX 1060 is used for testing. The recursion number $T$ of IM-blocks shown in Fig.~\ref{IMscheme} is determined by compromising between the dehazing performance and the processing time. As shown in Fig.~\ref{iteration}, increasing $T$ will boost the performance but increase the processing time, and vice versa. Finally, we choose $T_1 = T_2 = T_3 = 12$ as recursion number for each of three IM-blocks.

\subsection{Quantitative and Qualitative Evaluation}
We compare the proposed method with previous state-of-the-art methods, including DCP \cite{he2010single}, AOD-Net \cite{li2017aod},GFN \cite{ren2018gated}, DCPDN \cite{zhang2018densely}, GRID-Net \cite{liu2019griddehazenet}, FD-GAN \cite{dong2020fd}, and FFA-Net \cite{qin2019ffa}. We leverage the pre-trained models trained on RESIDE to reproduce the image comparison. In order to present our model's generality, all the aforementioned test datasets are evaluated using model trained on RESIDE training dataset. As shown in Tab.~\ref{performance}, our method outperforms previous methods except FFA on the SOTS synthetic dataset. 
Fig.~\ref{synthesis} shows the visual comparisons on the RESIDE, Fig.~\ref{real} shows the comparison on real hazy image.

\begin{figure*}[t]
    \noindent
    \centering
    \includegraphics[scale=0.34]{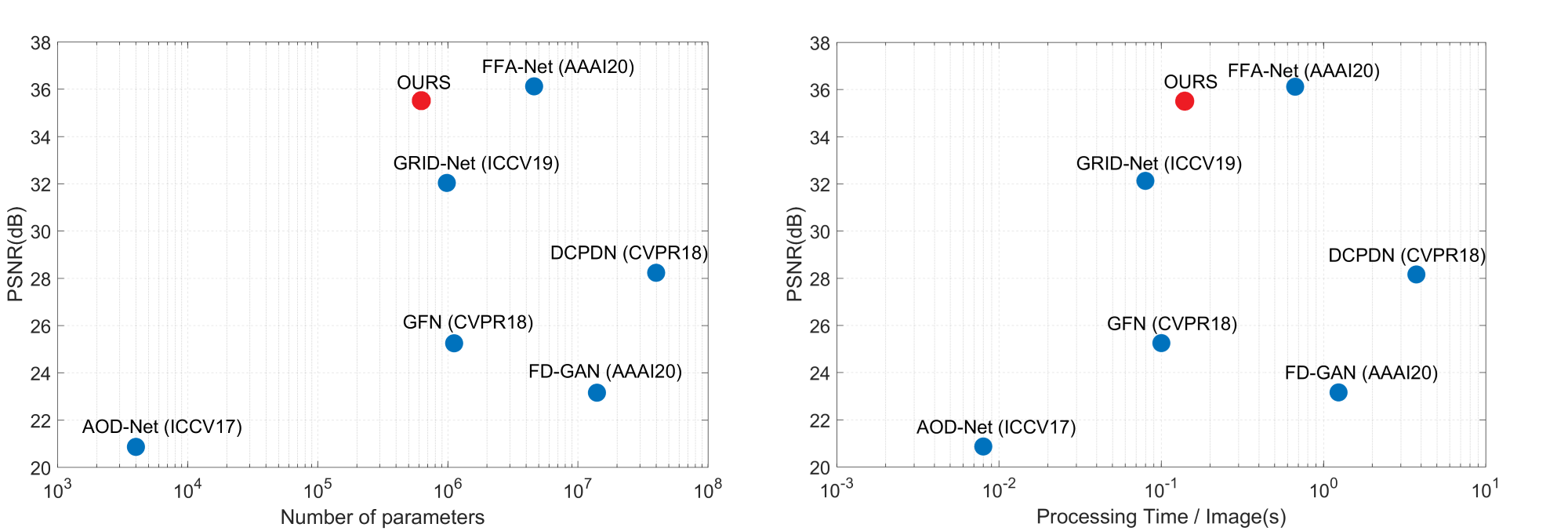}
    \caption{Performance of state-of-the-art methods versus the number of parameters on SOTS. The aforementioned methods are tested with NVIDIA GTX 1060 GPU. The results show that our work gets the best balance between performance and model size.}
    \label{efficiency}
\end{figure*}

The most of synthesis datasets \cite{li2018benchmarking} are based on atmospheric scattering model \cite{he2010single}, where depth information plays a crucial role. However, depth of transmission map is hard to measure from outside scene using depth telemeter. Without accurate depth information, outdoor synthesis hazy images are produced by setting depth information as a constant value. Compared to the outdoor synthesis hazing image, the indoor synthesis hazy image contains structure information due to accurate depth information measurement using depth telemeter. As shown in Fig.~\ref{out}, the IM-Net (ITS) performs much better on background area. Based on this observation, different from \cite{liu2019griddehazenet,zhang2018densely,qin2019ffa}, the quantitative comparisons of outdoor synthesis dataset are not presented in Table.~\ref{performance}.

Furthermore, in order to evaluate the computational efficiency of our proposed method to the aforementioned state-of-arts, we visualize the model size and running time per image versus performances with respect to PSNR on SOTS dataset, as shown in Fig.~\ref{efficiency}. Obviously, the results show that our proposed method gets the best balance between performance and model size, and thus can process very efficiently.

\begin{table}[t]
\centering
\caption{Ablation study on SOTS dataset.}
\label{ablation}
\setlength{\tabcolsep}{1mm}{
\begin{tabular}{c|cccccc}
\hline
\textbf{IM-block} &           &           & \checkmark & \checkmark & \checkmark & \checkmark \\
Resblock-1        & \checkmark &           &           &           &           &           \\
Resblock-T        &           & \checkmark &           &           &           &           \\ \hline
\textbf{RCA}     & \checkmark & \checkmark & \checkmark &           &           & \checkmark \\
OA                &           &           &           &           & \checkmark &           \\ \hline
\textbf{MLF}      & \checkmark & \checkmark &           & \checkmark & \checkmark & \checkmark \\ \hline \hline
PSNR (dB)         & 32.61     & 34.79     & 34.63     & 34.92     & 35.31     & 35.51     \\ \hline
\end{tabular}}
\end{table}
\subsection{Ablation Study}
To demonstrate the effectiveness of three mechanism referred in proposed methods section, we conduct ablation experiments to test the performance of model with and without specific component in MI-Net on SOTS dataset.

According to different scheme structure, we compare the implicit scheme structure, i.e. IM-block and the explicit scheme structure, ResNet \eqref{eq:resnet}. To develop an explicit structure, we directly replace the IM-block with Resblock, denoting Resblock-1 with -1 denoting only $1$ Resblock. Moreover, in order to conduct a fair comparison, we also replicate the Resblock for $T$ times (not shared) to achieve a comparable computational complexity as IM-block, denoting ResNet-T. 

From Tab.~\ref{ablation}, the results are improved by introducing RCA-block into framework, which verifies the effectiveness of RCA-block. Additionally, we have done an ablation experiment between ordinary attention block (OA-block) \cite{woo2018cbam} and our RCA-block, RCA-block's final PSNR increases 0.2dB compared to ordinary attention block.

Lastly, to demonstrate the function of multi-level fusion (MLF), we design a structure that directly fed $x_3$ into RCA-block without fusing multi-level features. The result indicates that combining these three mechanism can improve performance by a large margin.

\section{Conclusion}
We propose an end-to-end multi-level implicit network (MI-Net) for single image dehazing. The crucial idea of this work is to introduce a novel efficient implicit block that substitutes Resblock in image dehazing tasks. Moreover, MLF mechanism and RCA-block that modify the ordinary attention block are adopted to boost performance. Extensive experimental results demonstrate our method's superior performance over state-of-the-art methods.

\section*{Acknowledgement}
This work is supported by National Natural Science Foundation of China under Grant 61871297. We thank Jingwei He and Binyi Su for their insightful discussion about this work.

\newpage

{\small
\bibliographystyle{ieeetr}
\bibliography{ref}
}

\end{document}